\documentclass[final]{cvpr}

\usepackage{times}
\usepackage{epsfig}
\usepackage{graphicx}
\usepackage{amsmath}
\usepackage{amssymb}
\usepackage{nicefrac}       
\usepackage{microtype}      
\usepackage{xcolor}
\usepackage{grffile}
\usepackage{mwe}
\usepackage{verbatim}
\usepackage{setspace}
\usepackage{booktabs,arydshln}
\usepackage[format=plain,labelformat=simple,labelsep=period,font=small,skip=4pt,compatibility=false]{caption}
\usepackage[font=scriptsize,skip=2pt]{subcaption}
\usepackage{booktabs,tabularx}
\usepackage[inline]{enumitem}
\usepackage{mathtools}
\usepackage[ruled,linesnumbered]{algorithm2e}
\usepackage[capitalize]{cleveref}
\usepackage{adjustbox}
\usepackage{etoolbox,siunitx}
\robustify\bfseries
\robustify\itshape
\usepackage[keeplastbox]{flushend}
\usepackage{newtxtt}
\usepackage{newtxtt}
\usepackage{appendix}

\graphicspath{ {./Images/} }
\newcommand{\myparagraph}[1]{\smallskip\noindent\textbf{#1}}
\newcolumntype{C}[1]{>{\centering\let\newline\\\arraybackslash\hspace{0pt}}m{#1}}

\newcommand\Tstrut{\rule{0pt}{2.6ex}}         
\newcommand\Bstrut{\rule[-0.9ex]{0pt}{0pt}}   

\newcommand{\diff}{\mathop{}\!d}

\makeatletter
\def\adl@drawiv#1#2#3{%
        \hskip.5\tabcolsep
        \xleaders#3{#2.5\@tempdimb #1{1}#2.5\@tempdimb}%
                #2\z@ plus1fil minus1fil\relax
        \hskip.5\tabcolsep}
\newcommand{\cdashlinelr}[1]{%
  \noalign{\vskip\aboverulesep
           \global\let\@dashdrawstore\adl@draw
           \global\let\adl@draw\adl@drawiv}
  \cdashline{#1}
  \noalign{\global\let\adl@draw\@dashdrawstore
           \vskip\belowrulesep}}
\makeatother

\newcommand{\bernt}[1]{}
\newcommand{\mario}[1]{}
\newcommand{\mjf}[1]{}

\def\cvprPaperID{9937} 
\def\confYear{CVPR 2021}

\usepackage{fancyhdr}
\usepackage{setspace}

\begin{document}

\title{Euro-PVI: Pedestrian Vehicle Interactions in Dense Urban Centers}

\author{Apratim Bhattacharyya$^{1}$\quad Daniel Olmeda Reino$^2$  \quad  Mario Fritz$^{3}$ \quad Bernt Schiele$^1$ \\[0.5em]
$^1$Max Planck Institute for Informatics, Saarland Informatics Campus\\[0.10em]
$^2$Toyota Motor Europe \\[0.10em]
$^3$CISPA Helmholtz Center for Information Security\\
}

\maketitle


\begin{abstract}
Accurate prediction of pedestrian and bicyclist paths is integral to the development of reliable autonomous vehicles in dense urban environments. 
The interactions between vehicle and pedestrian or bicyclist have a significant impact on the trajectories of traffic participants e.g.\!\! stopping or turning to avoid collisions. 
Although recent datasets and trajectory prediction approaches have fostered the development of autonomous vehicles yet the amount of vehicle-pedestrian (bicyclist) interactions modeled are sparse.
In this work, we propose Euro-PVI, a dataset of pedestrian and bicyclist trajectories.
In particular, our dataset caters more diverse and complex interactions in dense urban scenarios compared to the existing datasets. 
To address the  challenges in predicting future trajectories with dense interactions, we develop a joint inference model that learns an expressive multi-modal shared latent space across agents in the urban scene. 
This enables our Joint-$\beta$-cVAE approach to better model the distribution of future trajectories.
We achieve state of the art results on the nuScenes and Euro-PVI datasets demonstrating the importance of capturing interactions between ego-vehicle and pedestrians (bicyclists) for accurate predictions.
\end{abstract}

\section{Introduction}
\begin{figure}[t]
    \centering
    
    \begin{tabularx}{\linewidth}{@{}XX@{}}
    \begin{subfigure}{\linewidth}\centering\includegraphics[width=\linewidth]{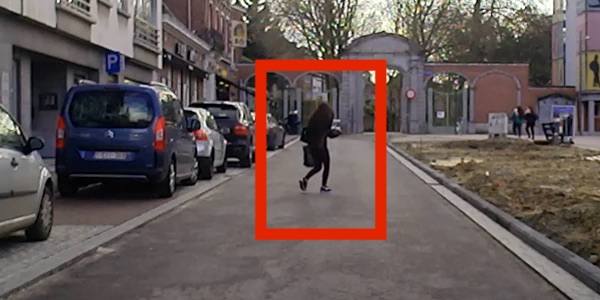}\caption{\scriptsize{Pedestrian \textbf{speeds} to avoid vehicle.}}\label{fig:teaser_a}\end{subfigure} &
    \begin{subfigure}{\linewidth}\centering\includegraphics[width=\linewidth]{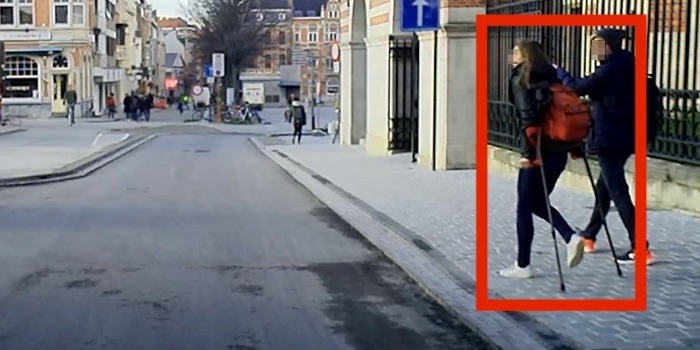}\caption{\scriptsize{\scriptsize{Pedestrians \textbf{yield} to the vehicle.}}}\label{fig:teaser_b}\end{subfigure} \\
    \begin{subfigure}{\linewidth}\vspace{0.5em}\centering\includegraphics[width=\linewidth]{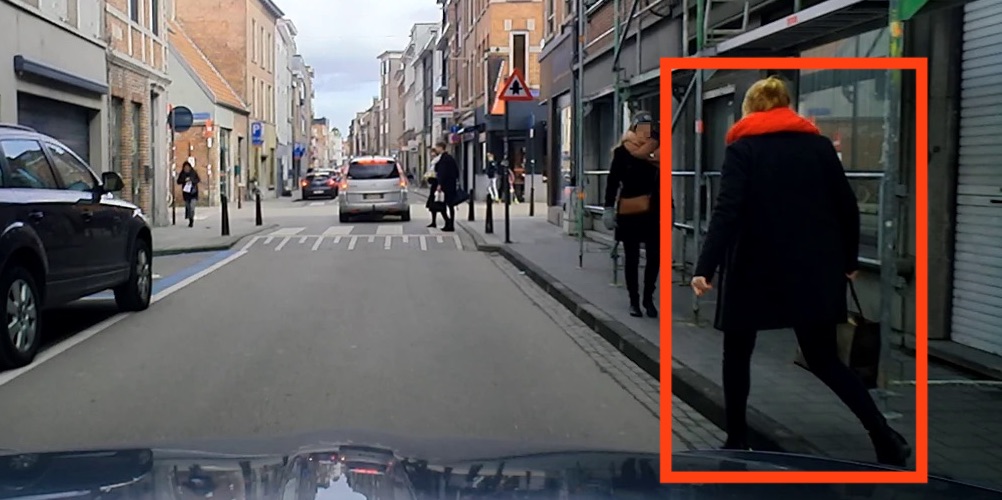}\caption{\scriptsize{Vehicle \textbf{slows} to avoid pedestrian.}}\label{fig:teaser_c}\end{subfigure} &
     \begin{subfigure}{\linewidth}\vspace{0.5em}\centering\includegraphics[width=\linewidth]{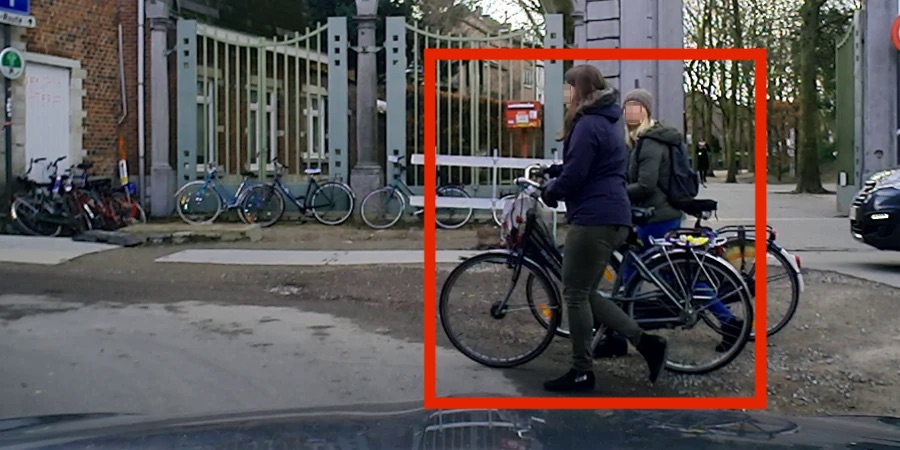}\caption{ \scriptsize{Vehicle \textbf{yields} to the bicyclists.}}\label{fig:teaser_d}\end{subfigure}\\
    \end{tabularx}
    
    \caption{Examples of interactions between the ego-vehicle and pedestrians (bicyclists) in Euro-PVI.}
    \label{fig:teaser}
\end{figure}

Notwithstanding recent progress in the development of reliable self-driving vehicles, dense inner city environments remain challenging. One of the key components for the success of self-driving vehicles in dense urban environments is anticipation \cite{BhattacharyyaSF18,LeeCVCTC17}. 
Anticipating the motion of traffic participants in dense urban environments is made especially challenging due to the effect of interactions between different agents. 
For example, a pedestrian might turn onto the road to avoid an obstacle on the sidewalk which requires the vehicle to stop (\cref{fig:teaser_c}). 
Alternately, a pedestrian attempting to cross the road ahead of the ego-vehicle might continue or stop depending upon the distance and velocity of the vehicle (\cf \cref{fig:example_interactions}). Thus, interactions add significant complexity to the distribution of the likely future trajectories which is highly multi-modal.

Recently, datasets like nuScenes \cite{CaesarBLVLXKPBB20}, Argoverse \cite{ChangLSSBHW0LRH19}, or Lyft L5 \cite{abs-2006-14480} have greatly aided the development of trajectory prediction methods. However, these datasets are primarily focused on trajectories of vehicles and vehicle-vehicle interactions -- collected to a large extent on multi-lane roads in North America or Asia, with sparse interactions between the ego-vehicle and pedestrians or bicyclists (\eg Figure 4 in \cite{abs-2006-14480}). Therefore, they do not represent trajectories in dense urban environments well where interactions between the trajectories of agents are prominent. Such scenarios are particularly common in inner-city environments in Europe.

In this work, we propose the new European Pedestrian Vehicle Interaction (\emph{Euro-PVI}) dataset \footnote{The dataset is available at \url{www.europvi.mpi-inf.mpg.de}} which is collected in a dense urban environment in Brussels and Leuven, Belgium.
The Euro-PVI dataset contains a rich and diverse set of interactions between the ego-vehicle and pedestrians (bicyclists). Sequences are recorded near busy urban landmarks \eg railway stations, narrow lanes or intersections (\cf \cref{fig:teaser,fig:map_teaser,fig:example_interactions}) where interactions are frequent and it is therefore challenging to predict pedestrian (bicyclist) paths.

Further, in spite of the recent progress in trajectory prediction methods, accurately capturing the multi-modal distribution of future trajectories \eg in dense urban environments remains challenging.
Current state of the art \cite{abs-1908-09008,MangalamGALAMG20,TrajTim20} generative models for trajectory prediction encode interactions directly in the posterior. 
Thus, the latent space does not express interaction information from the input distribution which limits the accuracy of the generated future trajectories. 
To address this limitation, we develop a latent variable deep generative model which jointly models the distribution of future trajectories of the agents in the scene. 
Our \emph{Joint-$\beta$-Conditional Variational Autoencoder (Joint-$\beta$-cVAE)} models a ``shared'' latent space between agents, to better capture the effect of interactions in the latent space and accurately represent the multi-modal distribution of trajectories.

Our contributions are, \begin{enumerate*}
    \item We propose Euro-PVI, a novel dataset of pedestrian and bicyclist trajectories recorded in Europe with dense interactions with the ego-vehicle.
    \item Our dataset facilitates research on dense interactions as we show that -- in contrast to prior datasets -- there is a pronounced performance gap between methods that model vehicle-pedestrian-interaction vs not.
    \item To this end, we develop a latent generative model  -- Joint-$\beta$-cVAE -- that models a shared latent space to better capture the effect of interactions on the multi-modal distribution of future trajectories.
    \item Finally, we demonstrate state of the art performance on pedestrian (bicyclist) trajectory prediction on nuScenes and Euro-PVI.
\end{enumerate*}

\begin{figure*}[t]
  \centering
  \begin{tabularx}{\textwidth}{@{}ccc@{}}
	\toprule
	\textbf{On-board Observation} & \textbf{$L_2$ Norm of Velocity} & \textbf{$L_2$ Norm of Acceleration} \\
	\midrule
	\includegraphics[width=0.3\linewidth,height= 2.7 cm]{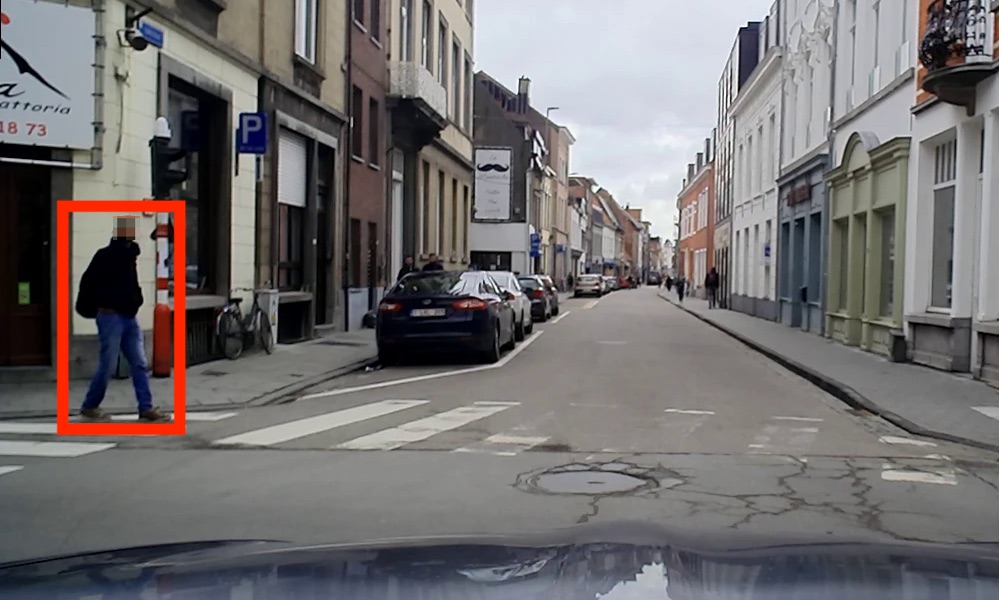}  &
	\includegraphics[width=0.32\linewidth,height= 2.7 cm]{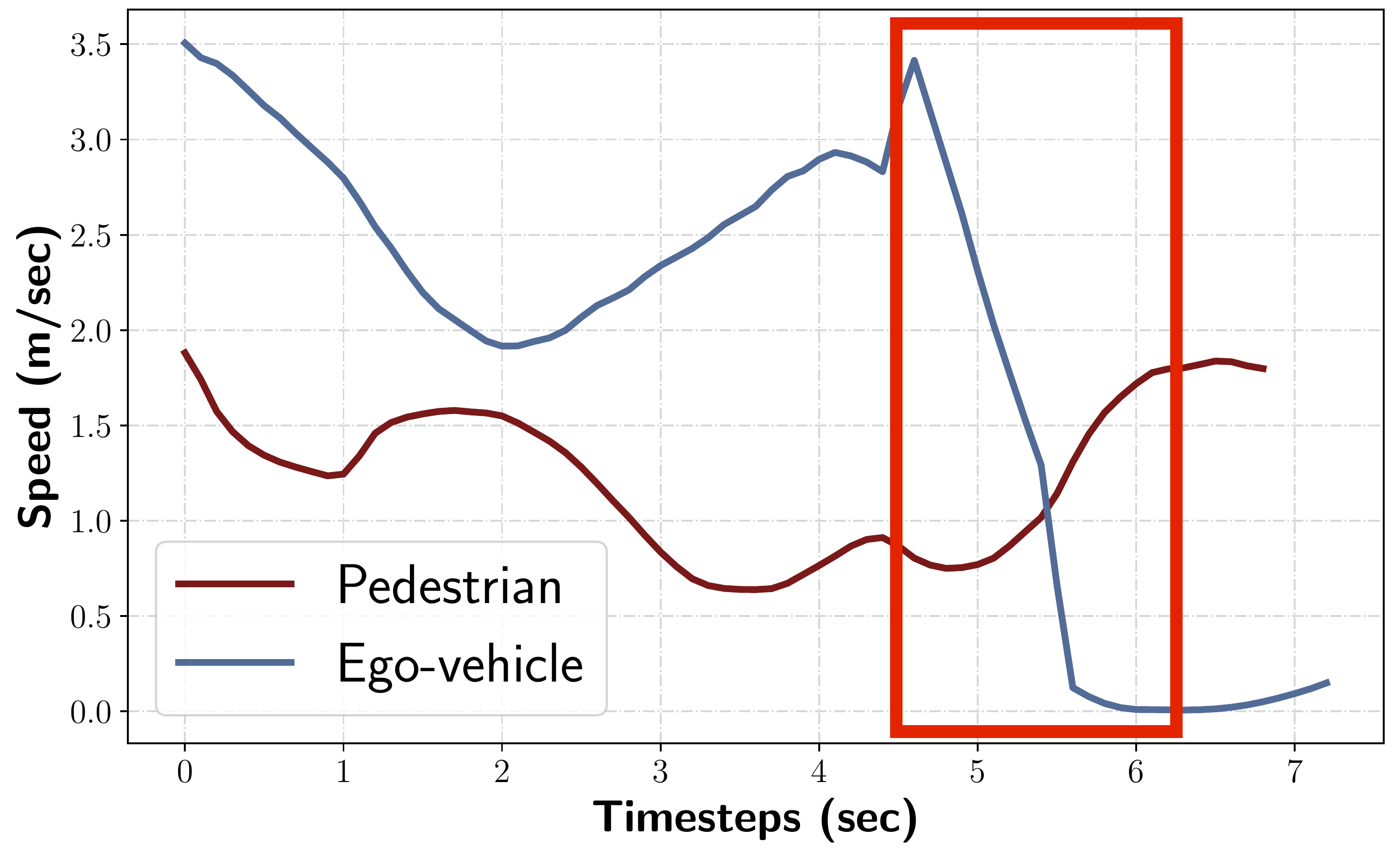} &
	\includegraphics[width=0.32\linewidth,height= 2.7 cm]{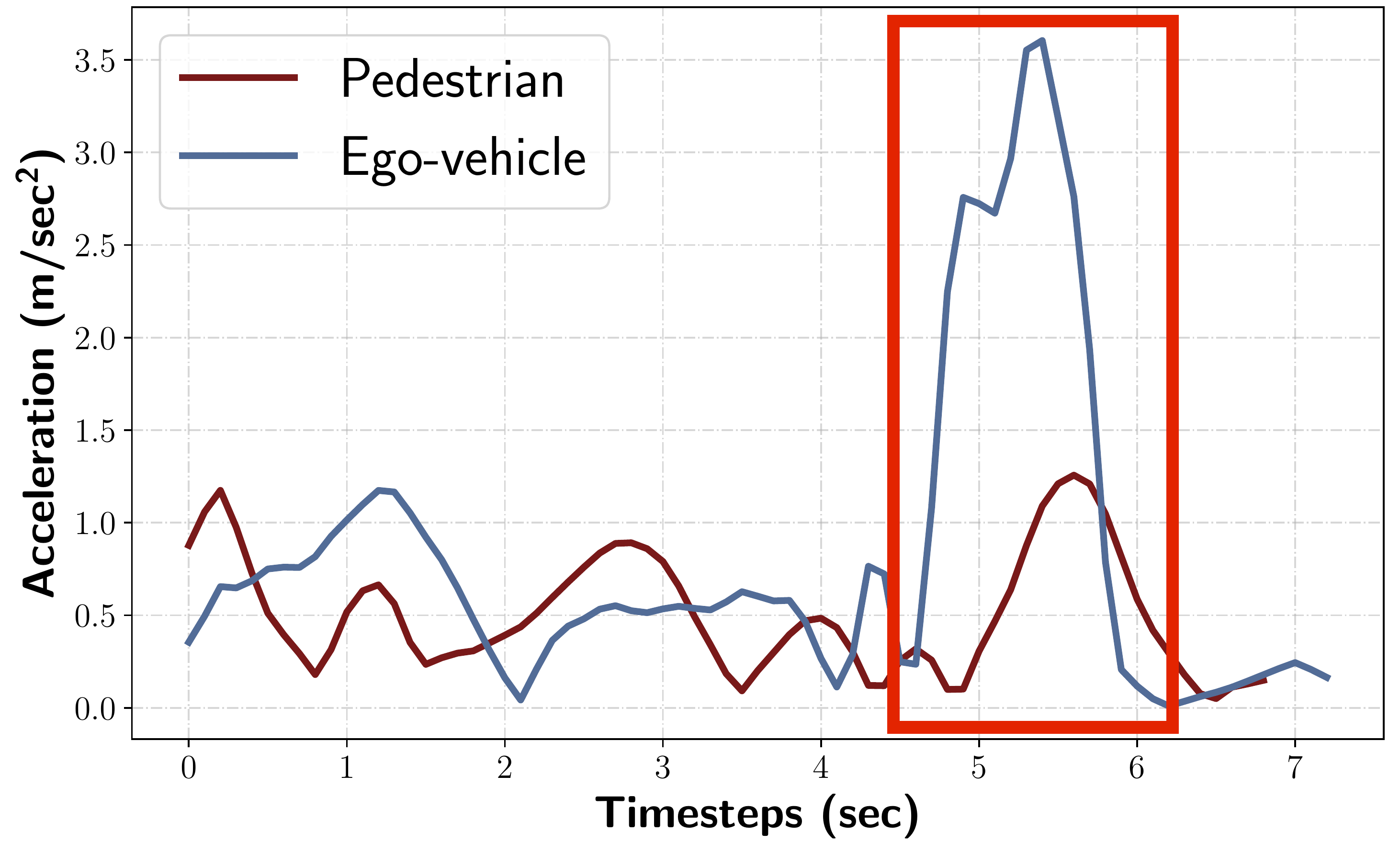}\\
	\cdashlinelr{1-3}
	\multicolumn{3}{c}{{\color[HTML]{791919} Pedestrian}: \textbf{First slows due to approaching vehicle, then crosses the street.}  {\color[HTML]{526C97}Ego-vehicle}: \textbf{Yields to pedestrian. }} \\
	\midrule
	\includegraphics[width=0.3\linewidth,height= 2.7 cm]{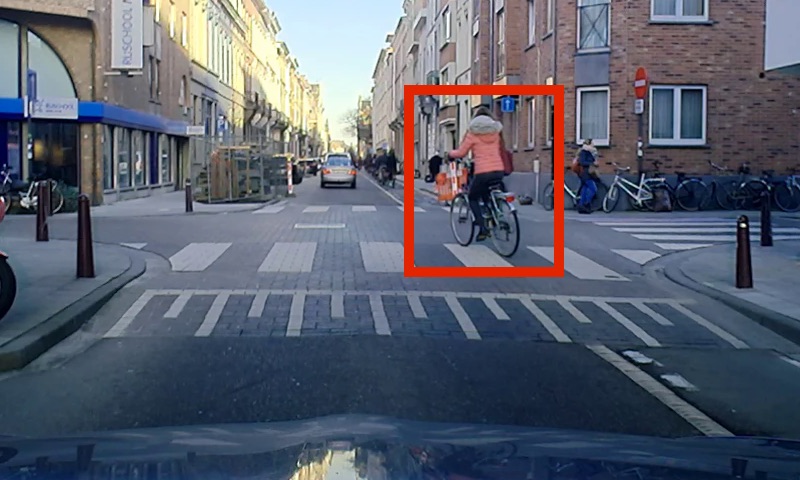}  &
	\includegraphics[width=0.32\linewidth,height= 2.7 cm]{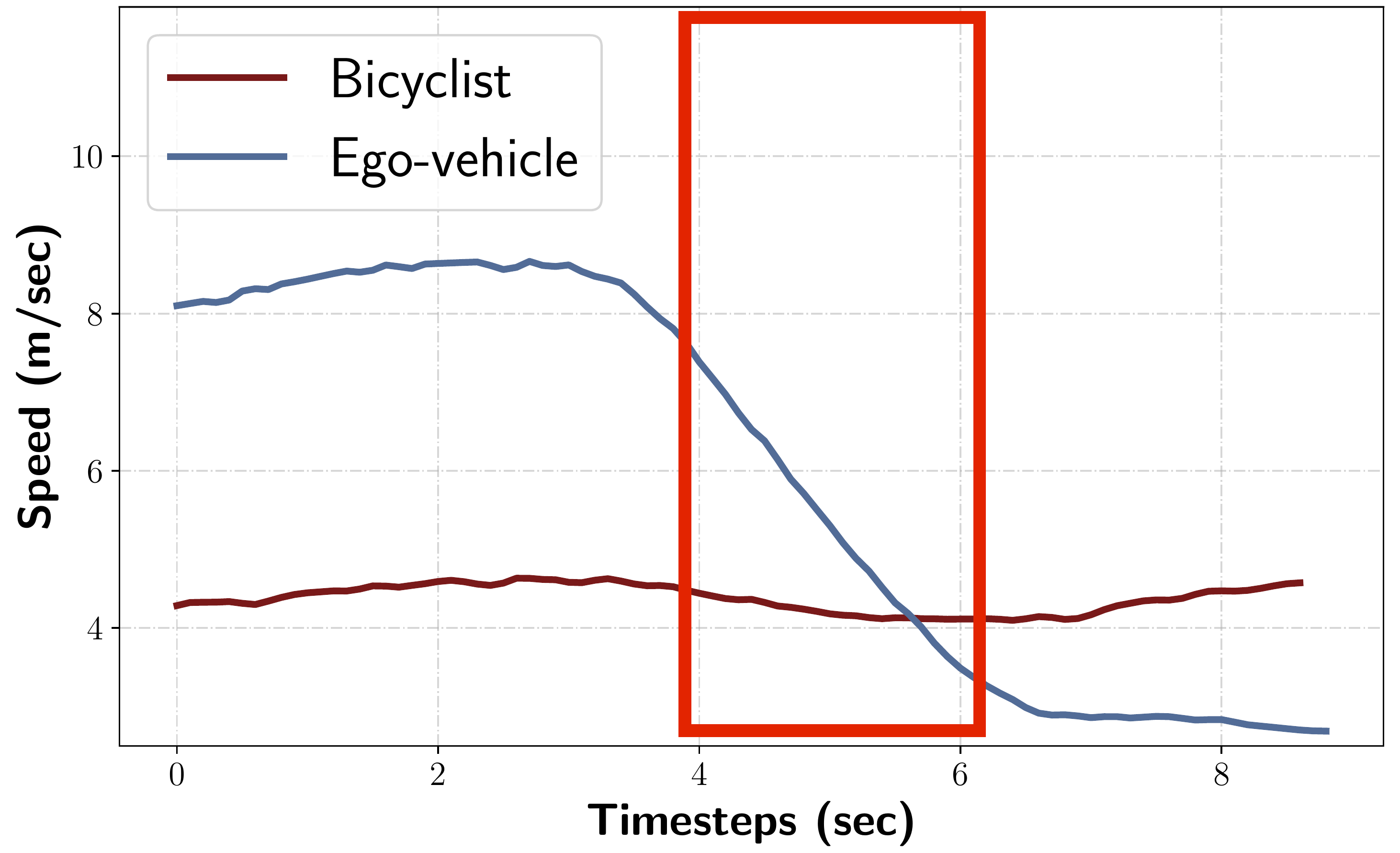} &
	\includegraphics[width=0.32\linewidth,height= 2.7 cm]{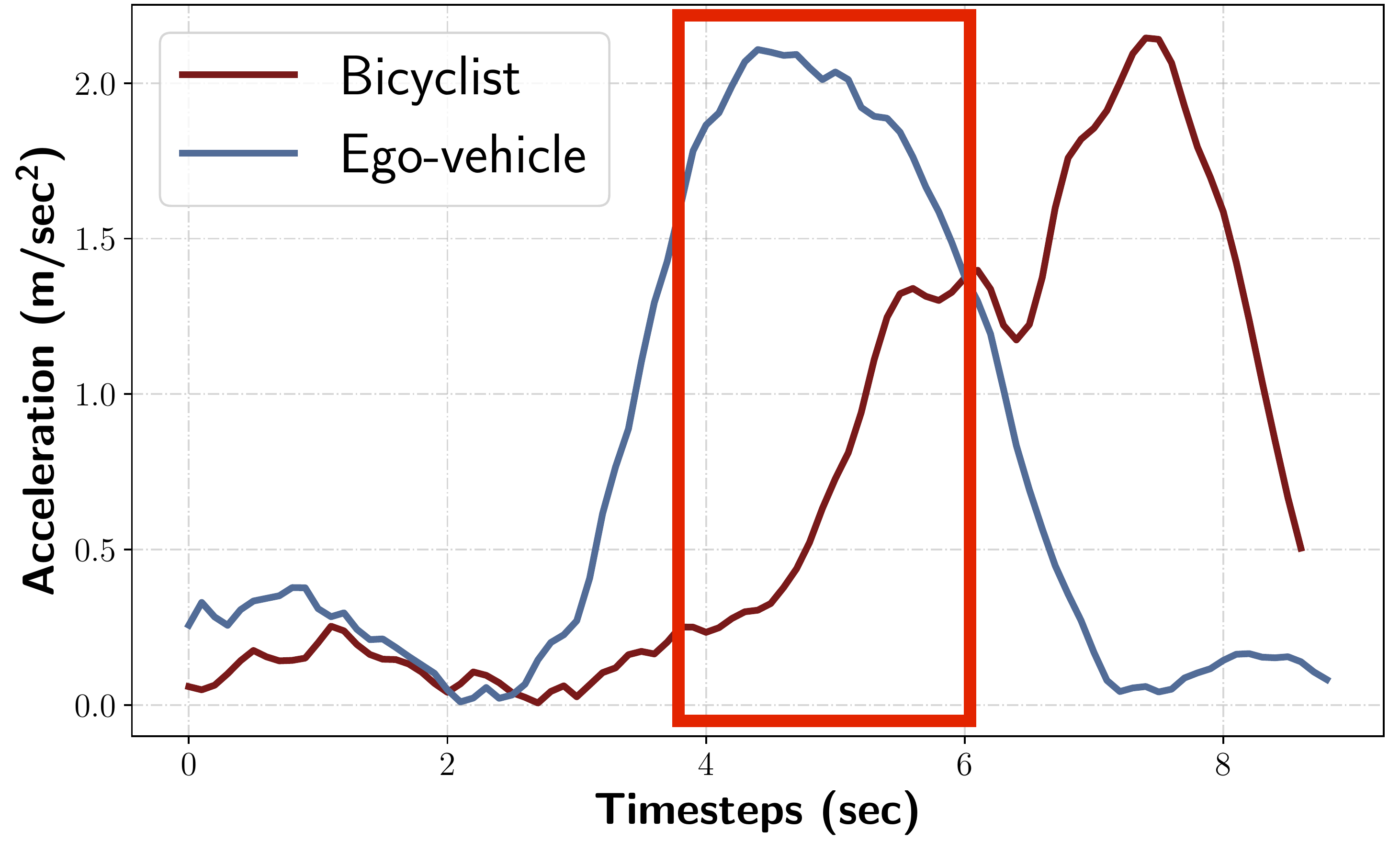}\\
	\cdashlinelr{1-3}
	\multicolumn{3}{c}{{\color[HTML]{791919}Bicyclist}: \textbf{Signals and turns left.} {\color[HTML]{526C97}Ego-vehicle}: \textbf{Slows down to avoid bicyclist.} } \\
	\bottomrule
  \end{tabularx}
  \caption{Examples of interactions in Euro-PVI. Spikes in the magnitude ($L_2$ norm) of acceleration resulting from interactions are marked. }
  \label{fig:example_interactions}
\end{figure*}
\section{Related Work}
\myparagraph{``On-board'' Trajectory Prediction Datasets.} Datasets like ETH/UCY \cite{LernerCL07} and Stanford Drone \cite{PellegriniESG09} are among the first datasets in the field of pedestrian trajectory prediction.
However, they are recorded using a birds eye view camera or drone. 
In order to aid the development of autonomous driving capabilities, recent datasets have moved to a more realistic ``on-board'' setting -- recorded from a (ego-)vehicle.
The popular ``on-board'' datasets: nuScenes \cite{CaesarBLVLXKPBB20}, Argoverse \cite{ChangLSSBHW0LRH19} and Lyft L5 \cite{abs-2006-14480} focus primarily on trajectories of nearby vehicles. 
nuScenes \cite{CaesarBLVLXKPBB20} and Argoverse \cite{ChangLSSBHW0LRH19} do not include annotated pedestrian (bicyclist) trajectories in their test set. 
Although Lyft L5 \cite{abs-2006-14480} includes pedestrian and bicyclist trajectories in the test set they are relatively rare (5.91\% and 1.62\% of all trajectories) as the chosen route does not include significant portions of dense urban environments. 
Moreover, in comparison to nuScenes \cite{CaesarBLVLXKPBB20} and Argoverse \cite{ChangLSSBHW0LRH19}, Lyft L5 \cite{abs-2006-14480} has lower diversity in terms of locations as it recorded only along a fixed route ($\sim\!6$km) in Palo Alto, California. 
Additionally, Lyft L5 \cite{abs-2006-14480} does not provide images from cameras and lidar point clouds, which are sources of rich contextual information. 
In contrast, PIE \cite{RasouliKKT19}, TITAN \cite{MallaDC20} and TRAF \cite{ChandraBBM19} focus primarily on pedestrians (bicyclists). 
However, the trajectories are recorded as sequences of 2D bounding boxes in the image plane and are not localized in the 3D world. 
The ApolloScapes \cite{MaZZYWM19} dataset does not include the trajectory of the ego-vehicle or contextual information \eg images from cameras or lidar point clouds.
Finally, note that these datasets are recorded either in North America or Asia and no large scale trajectory datasets are available for Europe. Euro-PVI is the first large scale dataset recorded in Europe dedicated to the task of trajectory prediction and unlike the existing datasets focuses on interactions between the ego-vehicle and  pedestrian (bicyclist).

\myparagraph{Methods for Trajectory Prediction.} Approaches which recognize the importance of interactions for the task of trajectory prediction date back to the Social Forces approach \cite{helbing1995social}. 
More recently Social-LSTM \cite{AlahiGRRLS16} proposes a social pooling mechanism to capture interactions. 
The social pooling mechanism has been extended in \cite{DeoT18} for efficiency. Alternately, TrafficPredict \cite{MaZZYWM19} uses an instance layer to embed a spatio-temporal graph of interactions. 
MATF \cite{ZhaoXMCBZ0W19} uses a tensor fusion scheme to capture interactions while retaining the spatial structure of the scene. 
TraPHic \cite{ChandraBBM19} proposes weighted interactions. 
Car-Net \cite{SadeghianLVVAS18} uses an attention scheme to integrate visual cues with interactions. 
Joint prediction of trajectories and activities is performed in \cite{Liang0NH019}. 
Social-STGCNN \cite{MohamedQEC20} models the prediction scene as a graph and uses graph convolutions to capture interactions. 
Spatio-temporal graph transformer networks are used in \cite{YuMRZY20}. 
A planning informed method is developed in \cite{SongDCSWC20}. 
However, these works largely assume a uni-modal future or a future conditioned using a set of possible maneuvers and cannot fully capture multi-modal distributions \eg trajectories in dense urban environments.

Therefore, recent work has focused on the development of generative approaches to capture multi-modal distributions of trajectories, primarily using conditional GAN \cite{MirzaO14}, conditional Normalizing flows \cite{DinhSB17} or conditional VAE \cite{SohnLY15} based models. 
Social-GAN \cite{GuptaJFSA18} proposes to use social pooling in a conditional GAN setup to capture the effect of social interactions in the distribution of future pedestrian trajectories. 
Sophie \cite{SadeghianKSHRS19} uses an attention mechanism in a conditional GAN setup to capture the effect of interactions. 
Graph attention networks are employed by \cite{KosarajuSM0RS19} in a conditional GAN setup. 
Normalizing flows are combined with a context attention mechanism in \cite{abs-2003-03212} to improve expressiveness in the latent space. 
HBA-Flow \cite{abs-2009-09878} improves trajectory prediction with a Haar-wavelet based decomposition.
DESIRE \cite{LeeCVCTC17} uses a RNN refinement module over the plain cVAE setup. Predictions that ``personalize'' to agent behaviour are proposed in \cite{FelsenLG18}. 
The standard cVAE objective is improved in \cite{BhattacharyyaSF18} for better match between the predicted and true distributions. 
A social graph network is used to condition a cVAE in \cite{abs-1907-10233}. It is shown in \cite{MangalamGALAMG20} that conditioning the cVAE additionally on the goal state can significantly improve accuracy. 
Finally, Trajectron++ \cite{TrajTim20} uses a pooled scene graph in a cVAE framework to capture interactions. 
The resulting approach leads to state of the art pedestrian trajectory prediction both in the ETH/UCY datasets and in the ``on-board'' setting of nuScenes. 
Although cVAE based methods achieve state of the art performance, the latent space in the above mentioned works does not encode the effect of interactions. 
This makes it challenging to accurately capture the trajectory distribution \eg in dense urban environments where these interactions have a significant effect. 
In this work, we propose to use a shared latent space across agents to better capture the effect of interactions to generate accurate future trajectories.

\section{The Euro-PVI Dataset}
\label{sec:Eurp-PVI D}
In this section, we introduce our Euro-PVI dataset to advance the task of ``on-board'' trajectory prediction especially in dense urban environments. The dataset focuses on the role of the ego-vehicle - pedestrian (bicyclist) interactions present in a scene to predict the future trajectories in dense urban environments. We first concretely define an ``interaction'', which guides our data collection process and helps us select relevant sequences. Next, we provide details of the sensor setup and the data collection process. We then compare Euro-PVI to the existing datasets with respect to the density of interactions and provide detailed dataset statistics.

\myparagraph{Interactions.} We define an interaction between the ego-vehicle and a traffic participant (\eg pedestrian or bicyclist) as an event where the presence of either (or both) the ego-vehicle or the traffic participant causes a change in velocity (change of speed/direction of motion \ie acceleration) of the other. Examples of interactions include, \begin{enumerate*}
    \item The ego-vehicle yielding to a pedestrian at a crosswalk (\cref{fig:example_interactions} top).
    \item Non-verbal communication causes the ego-vehicle to slow down, as to avoid a bicycle which wants to turn (\cref{fig:example_interactions} bottom).
\end{enumerate*} 
We aim to record sequences which contain dense interactions between the ego-vehicle and vulnerable traffic participants, in particular, pedestrians and bicyclists.

\begin{figure}[t]
     \centering
     \begin{subfigure}[b]{0.22\textwidth}
         \centering
         \includegraphics[width=\textwidth]{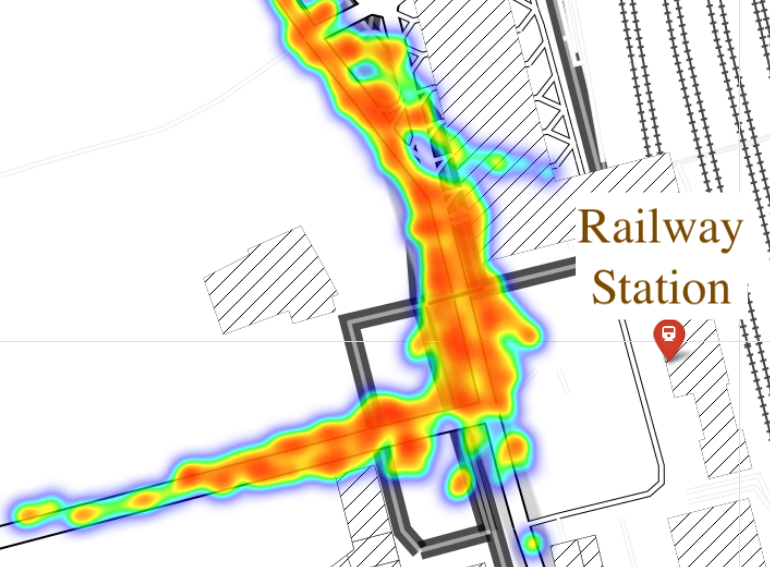}
     \end{subfigure}
     \begin{subfigure}[b]{0.22\textwidth}
         \centering
         \includegraphics[width=\textwidth]{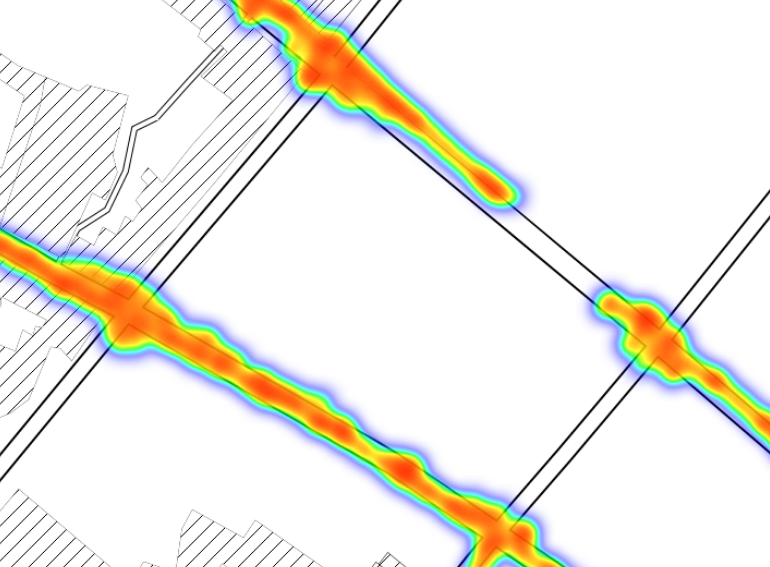}
     \end{subfigure}
        
    \caption{Examples of aggregated spatial distribution of trajectories of pedestrians and cyclists around intersections and urban landmarks.}
    \label{fig:map_teaser}
\end{figure}

\myparagraph{Drive Planning and Scene Selection.} Euro-PVI is recorded in the dense urban scene of Brussels and Leuven, Belgium. The ego-vehicle is equipped with a $360^{\circ}$ lidar, a positioning system and a set of front-facing synchronized cameras (see appendix). The ego-vehicle is crewed by a pilot and a co-pilot. The pilot is instructed to drive freely over a predetermined area and to re-visit locations at times when dense concentration of pedestrians (bicyclists) is to be expected, such as transport hubs during peak hour. 
The duration of the driving sessions were up to 8 hours per day, over the course of two weeks. The co-pilot is tasked with identifying interactions and tagging the event. 
In the case of changes in trajectory or velocity of the ego-vehicle, the co-pilot asks the pilot for confirmation. 
The pilot is also instructed to spontaneously indicate that an interaction has happened. 
In \cref{fig:map_teaser} we show the geographical distributions of the trajectories in two example locations. We see that there is a high density of trajectories located around busy urban landmarks \eg the railway station. We also show that interactions are not confined to road intersections with crosswalks where pedestrian (bicyclist) trajectories are simpler to predict, but also occur at locations without crosswalks (\ie without designated crossing areas for  pedestrians/bicyclists) where trajectories are more challenging to predict.

\begin{figure}[t]
\centering
\begin{subfigure}[t]{.495\linewidth}
  \centering
  \includegraphics[height=2.85cm]{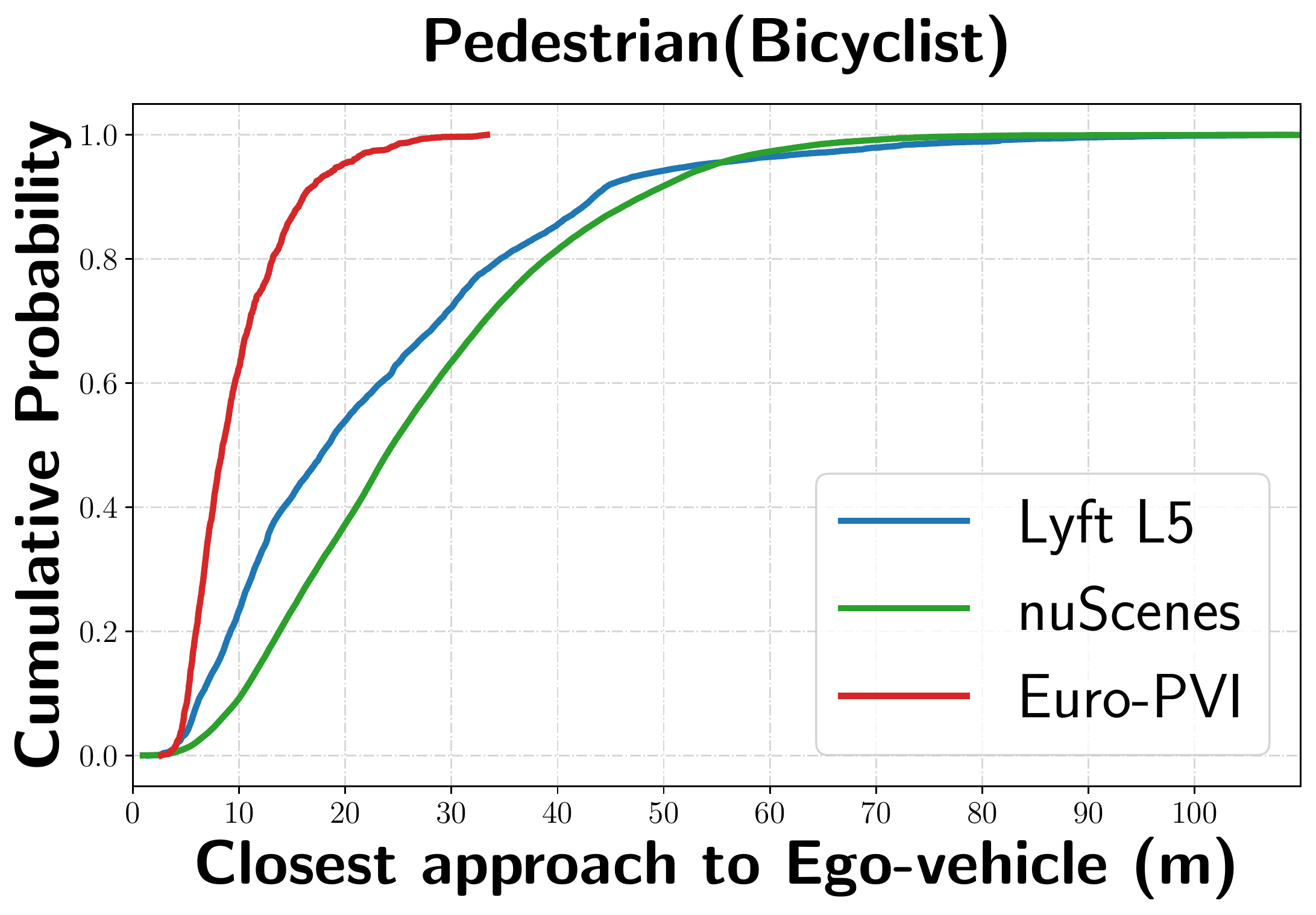}
\end{subfigure}%
\hfill
\begin{subfigure}[t]{.495\linewidth}
  \centering
  \includegraphics[height=2.85cm]{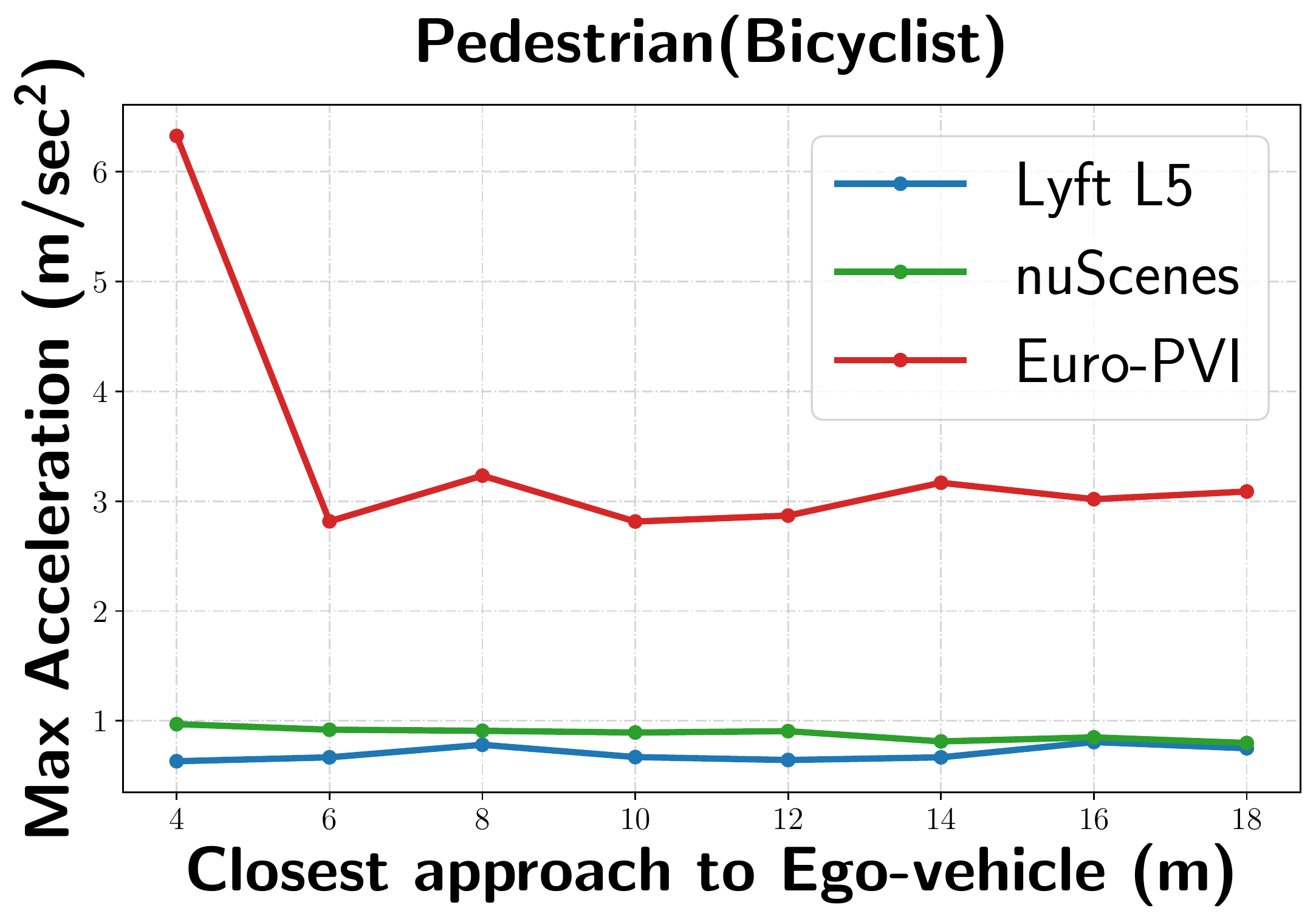}
\end{subfigure}%
\hfill
\caption{Left: Cumulative distribution sorted by distance to the ego-vehicle. Close proximity of the ego-vehicle and pedestrians(bicyclists) in Euro-PVI indicate dense traffic scenarios where interactions are likely. Right: Maximum acceleration sorted by distance to the ego-vehicle. High acceleration in close proximity of the ego-vehicle and pedestrians(bicyclists) indicate high likelihood of interactions. }
\label{fig:closestapproach}
\end{figure}

\myparagraph{Interactions in Urban Environments.} We now compare Euro-PVI to existing datasets for trajectory prediction with respect to the density of interactions, in particular to the two largest datasets -- nuScenes \cite{CaesarBLVLXKPBB20} and Lyft L5 \cite{abs-2006-14480}.
First, we compare the distances between the ego-vehicle and pedestrians (bicyclists) in the scene. Short distances are indicative of closely packed urban environments where interactions frequently occur. In \cref{fig:closestapproach} (left), we show the closest approach (proximity) of a pedestrian or bicyclist to the ego-vehicle.
We see that in the nuScenes and Lyft L5 datasets, the majority of the pedestrians (bicyclists) do not approach the ego-vehicle closer than $\sim\!20$ meters. Such large distances between the pedestrians (bicyclists) and the ego-vehicle, more than 4 typical car lengths, decreases the likelihood of interactions. 
In contrast, in Euro-PVI the majority of the pedestrians (bicyclists) approach the ego-vehicle is as close as $\sim\!8$ meters. 
Such short distances between the ego-vehicle and pedestrians (bicyclists) are indicative of the densely packed urban environment in which Euro-PVI is recorded with increases the likelihood of interactions. 

However, close proximity only increases the likelihood of interactions, but does not necessarily lead to interactions. 
Note that by definition, interactions lead to a change in velocity \ie acceleration. 
Therefore, in \cref{fig:closestapproach} (right) we plot the maximum acceleration experienced by the pedestrian (bicyclist) and ego-vehicle with increasing distance to the ego-vehicle and a pedestrian (bicyclist) respectively. In case of nuScenes and Lyft L5 we do not see a strong dependence on distance. 
In contrast, in Euro-PVI both pedestrians (bicyclists) and the ego-vehicle experience the maximum acceleration close to the point of closest approach. This is again strongly indicative of dense interactions in Euro-PVI. 
\begin{table*}[t]
  \centering
  \begin{adjustbox}{max width=\linewidth}
  \begin{tabular}{@{\extracolsep{3pt}}lccccccccc}
	\toprule
	& & & &  \multicolumn{2}{c}{\textbf{Trajectory Instances}} & \multicolumn{3}{c}{\textbf{Labels}} \\
	\cline{5-6} \cline{7-10}
	\textbf{Dataset} & \textbf{Location} & \textbf{Scenes} & \textbf{Length (hrs)} & \textbf{Pedestrians} & \textbf{Bicyclists}
	& \textbf{Front Camera} & \textbf{Lidar} & \textbf{Seg. Maps} & \textbf{IMU} \Tstrut\Bstrut\\
	\midrule
	nuScenes \cite{CaesarBLVLXKPBB20} & North Am. \& Asia & 1000 & 5.5 & 9142 & 550 & \checkmark & \checkmark & \checkmark & \checkmark \\
	ApolloScapes \cite{MaZZYWM19} & Asia & 54 & 1.7 & 3065 & 1827\textsuperscript{ *} & x & x & x & x \\
	Lyft L5 \cite{abs-2006-14480} & North Am. & \emph{170$\times$10\textsuperscript{ 3}} & \emph{1118} & \emph{605$\times$10\textsuperscript{ 3}} & \emph{77$\times$10\textsuperscript{ 3}} & x & x & x & \checkmark\\
	\midrule 
	CityScapes \cite{CordtsORREBFRS16} & Europe & {5000} & {2.5} & 0 & 0 & \checkmark & x & \checkmark & \checkmark \\
	KITTI \cite{GeigerLU12} & Europe & 71 & 1.5 & 380 & 150 & \checkmark & \checkmark & \checkmark & \checkmark\\
	KITTI-360 \cite{XieKSG16} & Europe & 9 & {2.2} & 262 & 82 & \checkmark & \checkmark & \checkmark & \checkmark\\
	A2D2 \cite{abs-2004-06320} & Europe & 23 & 0.9 & 260 & 248 & \checkmark & \checkmark & \checkmark & \checkmark\\
	Euro-PVI (Ours) & Europe & {1077} & {2.2} & {6177} & {1581} & \checkmark & \checkmark & \checkmark & \checkmark \\
	\bottomrule
  \end{tabular}
  \end{adjustbox}
  \caption{Comparison of dataset statistics. (Seg. Maps -- semantic and/or instance segmentation, * includes motorbikes.)}
  \label{tab:stats}
\end{table*}

\myparagraph{Qualitative Examples of Interactions.} We provide example interactions in Euro-PVI along with the resultant acceleration of the involved agents in \cref{fig:example_interactions}, \eg top row: the pedestrian first slows down due to the approaching ego-vehicle and at the same time, the ego-vehicle sees the pedestrian and yields. This is visible as a spike in the velocity and acceleration plots. Similar spikes in acceleration can be observed due to interactions in the other examples in \cref{fig:example_interactions}. We provide additional examples in the appendix.

\myparagraph{Additional Statistics.} We report dataset statistics and available labels of Euro-PVI in \cref{tab:stats}. In addition to the annotated pedestrian (bicyclist) trajectories, Euro-PVI contains 83k camera images and the corresponding lidar point clouds along with synchronized IMU data. In terms of size Euro-PVI is competitive with nuScenes \cite{CaesarBLVLXKPBB20} and ApolloSpaces \cite{MaZZYWM19} and while Lyft L5 \cite{abs-2006-14480} is significantly larger, it does not provide labels \eg camera images or lidar point clouds. 
Furthermore, Euro-PVI surpasses the the largest autonomous driving datasets collected in Europe \ie CityScapes \cite{CordtsORREBFRS16}, KITTI \cite{GeigerLU12},  KITTI-360 \cite{XieKSG16} and A2D2 \cite{abs-2004-06320} in terms of number of instances of pedestrian (bicyclist) trajectories. 
CityScapes \cite{CordtsORREBFRS16} and A2D2 \cite{abs-2004-06320}, do not provide annotated 3D pedestrian or bicycle trajectories. KITTI \cite{GeigerLU12} and  KITTI-360 \cite{XieKSG16} contains mostly linear motion with sparse interactions and thus commonly not used for trajectory prediction \cite{MaZZYWM19,RhinehartKV18}. 
In fact, Euro-PVI is the first large scale dataset with dense interactions (\cref{fig:closestapproach}) dedicated to trajectory prediction in Europe, to the best of our knowledge.

\section{Joint-$\beta$-cVAE: A Joint Trajectory Model for Dense Urban Environments}
Following the observations in \cref{sec:Eurp-PVI D}, we find that vehicle-pedestrian (bicyclist) interactions are crucial to the task of future trajectory prediction in dense urban scenarios. 
In particular, inherent multi-modality of the distribution of future trajectories and the effect of interactions on this complex distribution, make accurately predicting the future trajectories in dense urban environments challenging.

Specifically, given a scene with $n$ agents \eg vehicles, pedestrians or bicyclists in a dense urban environment and the past observations $\mathbf{x}_i \in \mathbf{X}$ for each agent $i$, we model the future trajectories $\mathbf{y}_{i} \in \mathbf{Y}$ for each agent $i \in \{1, \ldots, n\}$ in the scene. 
Here the past observations $\mathbf{x}_i$ include the past trajectories and the past context corresponding to the past trajectory sequence.
Prior work \cite{abs-1908-09008,BhattacharyyaSF18,MangalamGALAMG20,TrajTim20} models the conditional distribution  $p_{\theta}(\mathbf{y}_{i} | \mathbf{X} )$ parameterized by $\theta$  of the future trajectories $\mathbf{y}_i$ using the latent variables $\mathbf{z}_i$ in the standard conditional VAE formulation \cite{HigginsMPBGBML17,KingmaW13,SohnLY15},
\begin{align}\label{eq:cvae}
    p_{\theta}(\mathbf{y}_{i} | \mathbf{X} ) =& \int p_{\theta}(\mathbf{y}_{i} | \mathbf{z}_{i}, \mathbf{X} ) \, p_{\theta}(\mathbf{z}_{i} | \mathbf{x}_{i} ) \diff \mathbf{z}_{i}.
\end{align}
Here, the distribution $p_{\theta}(\mathbf{z}_{i} | \mathbf{x}_{i} )$ assumes conditional independence of the latent variables $\{ \mathbf{z}_1, \ldots,\mathbf{z}_n \} \in \mathbf{Z} $ given the past observation $\mathbf{x}_i$ of each agent. This assumption essentially ignores the motion patterns of interacting agents \ie the ego-vehicle and other pedestrians (bicyclists) in the scene, which in real world dense urban scenarios is critical for the accurate prediction of future trajectories.
The formulation in \cref{eq:cvae} therefore limits the amount of interactions between the agents that can be encoded in the latent space.
Since, the latent variables $\mathbf{z}_i$ are crucial for capturing diverse futures in such conditional models, it is important to express the effect of interactions in the latent space.

We now introduce our Joint-$\beta$-cVAE approach which aims to accurately capture the effect of interactions in the latent space for trajectory prediction in dense urban environments. Our proposed Joint-$\beta$-cVAE model, in contrast to prior conditional VAE based models \cite{abs-1908-09008,BhattacharyyaSF18,MangalamGALAMG20,TrajTim20} encodes the joint distribution of the latent variables across all agents in the scene. This allows our Joint-$\beta$-cVAE model to encode the dependence of the future trajectory distribution on interacting agents in the latent space, leading to more accurate modeling of the multi-modal future trajectory distribution.

\begin{figure}[t]
\centering
  \includegraphics[height=5.5cm]{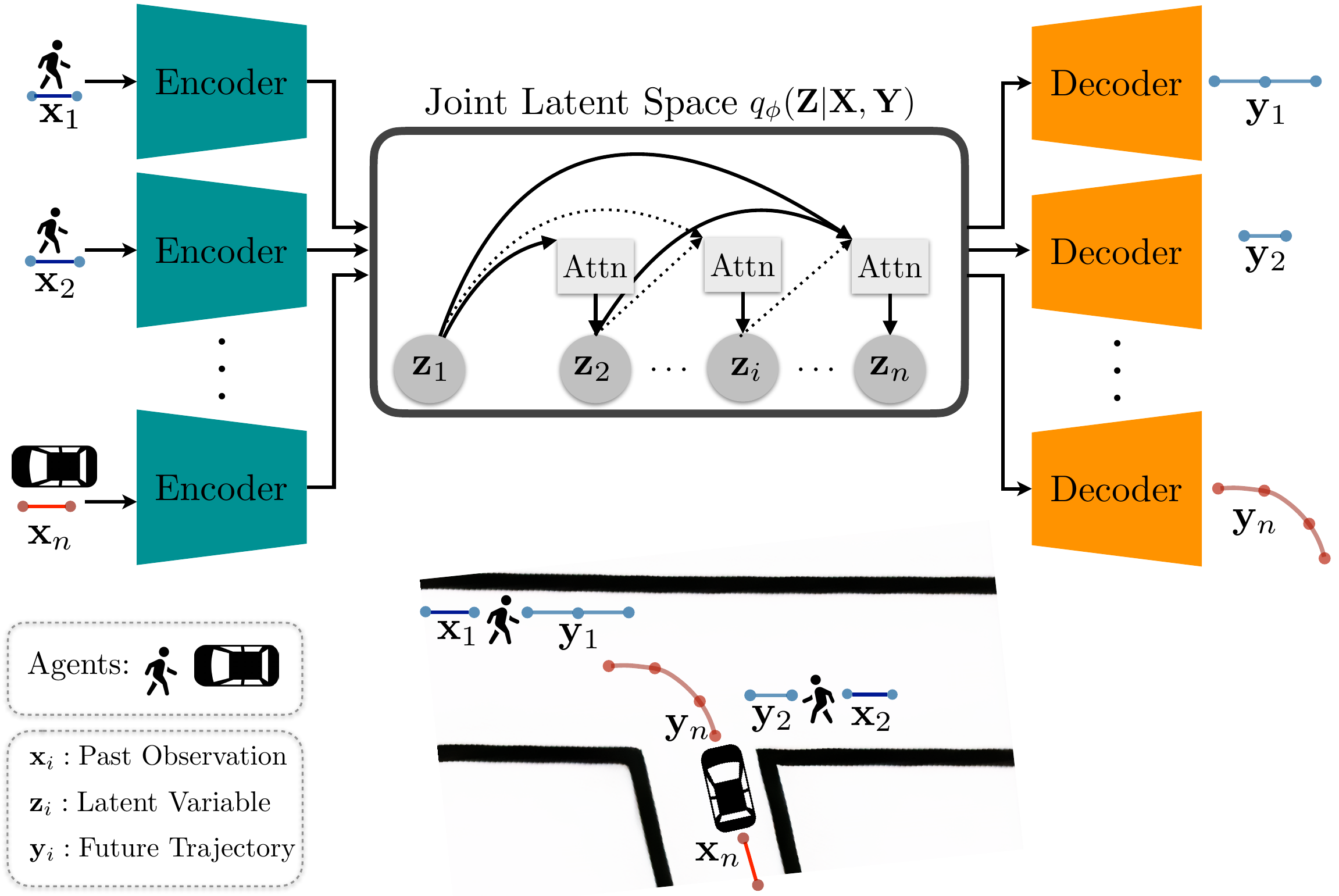}
\caption{Our Joint-$\beta$-cVAE, which models a Joint latent space across all agents $\{1,\cdots,n\}$ in the scene. The posterior latent distribution $q_{\phi}(\mathbf{Z} | \mathbf{X},\mathbf{Y})$ factorizes auto-regressively and models the dependence of $\mathbf{z}_i$ on $\{\mathbf{Z}_{< i},\mathbf{X},\mathbf{Y}\}$ using an attention mechanism.}
\label{fig:jointcvae}
\end{figure}

\myparagraph{Formulation.}  Our Joint-$\beta$-cVAE in \cref{fig:jointcvae} models the joint distribution of future trajectories $\mathbf{Y}$ across all $n$ agents, using the latent variables $\mathbf{z}_{i} \in \mathbf{Z}$,
\begin{align}\label{eq:jcvae}\raisetag{+1.7cm}
    \begin{split}
        p_{\theta}&(\mathbf{Y} | \mathbf{X} ) = \int p_{\theta}(\mathbf{Y}, \mathbf{Z} | \mathbf{X} ) \diff \mathbf{Z} \\
                                             =& \int \prod\limits_i^n p_{\theta}(\mathbf{y}_{i}, \mathbf{z}_{i} | \mathbf{Z}_{< i}, \mathbf{Y}_{< i}, \mathbf{X} ) \diff \mathbf{Z} \\
                                            =&\int \prod\limits_i^n p_{\theta}(\mathbf{y}_{i} | \mathbf{Z}_{\leq i}, \mathbf{Y}_{< i}, \mathbf{X} ) p_{\theta}( \mathbf{z}_i | \mathbf{Z}_{< i}, \mathbf{Y}_{< i}, \mathbf{X})  \diff \mathbf{Z} \\
    \end{split}
\end{align}
where $p_{\theta}(\mathbf{Y}, \mathbf{Z} | \mathbf{X} )$ is the joint distribution of the future trajectories and the latent variables for all agents. In the second step, without loss of generality, we auto-regressively factorize the joint distribution over the $n$ agents, where $\mathbf{Z}_{\leq i}, \mathbf{Y}_{< i}$ denotes the latent variables and trajectories for agents $\{1,\dots,i - 1\}$.
Note that, the factorization is agnostic to the choice of ordering of the agents. 
In contrast to \cref{eq:cvae}, the prior distribution of the latent variables in \cref{eq:jcvae} exhibits joint modeling of the latent variables $\mathbf{z}_i$, \ie $p_{\theta}( \mathbf{z}_i | \mathbf{Z}_{< i}, \mathbf{Y}_{< i}, \mathbf{X})$.

We maximize the log-likelihood of the data under the model in \cref{eq:jcvae} with variational inference using a joint variational posterior distribution $q_{\phi}(\mathbf{Z} | \mathbf{X}, \mathbf{Y})$.

\myparagraph{The Joint Posterior.} To encode rich latent spaces shared between the $n$ agents which capture the effect of interactions in dense urban environments, we propose a joint posterior over all $n$ agents in the scene, which auto-regressively factorizes,
\begin{align}\label{eq:jcvae_posterior}
    \begin{split}
        q_{\phi}(\mathbf{Z} | \mathbf{X}, \mathbf{Y} ) = q_{\phi}(\mathbf{z}_{1}  | \mathbf{X}, \mathbf{Y} ) \cdots 
        q_{\phi}(\mathbf{z}_{n}  | \mathbf{Z}_{< n},  \mathbf{X}, \mathbf{Y} ).
    \end{split}
\end{align}

The conditional distributions $q_{\phi}(\mathbf{z}_{i}  | \mathbf{Z}_{< i},  \mathbf{X},  \mathbf{Y} )$ corresponding to agent $i$ are normal distributions whose mean and variances are functions of  $\mathbf{Z}_{< i}$, $\mathbf{X}$ and $\mathbf{Y}$. 
Intuitively, given the past trajectories of the agents $\{1,\ldots,n\}$ and the joint latent posterior  distribution over the interacting agents $\{1,\ldots,i\}$, the latent posterior distribution corresponding to agent $i$ encoding its interactions can be inferred conditioned on the latent information of the interacting agents $\{1,\ldots,i-1\}$. We use an attention mechanism inspired by \cite{00010BT0GZ18}, to model the dependence of the distribution of $\mathbf{z}_i$ on $\mathbf{Z}_{< i}$ and $\{\mathbf{X},\mathbf{Y}\}$. The attention weights on $\mathbf{z}_j$ and $\{\mathbf{x}_j,\mathbf{y}_j\}$, for $j \neq i$, is additionally conditioned on the past observation and location of the agents (\cref{fig:jointcvae}) -- which allows to attend to agents $j$ interacting with agent $i$. 

In contrast, prior work \cite{abs-1908-09008,BhattacharyyaSF18,MangalamGALAMG20,TrajTim20} employ conditionally independent posteriors $q_{\phi}(\mathbf{z}_{i}  | \mathbf{x}_{i},  \mathbf{y}_i )$ across agents in a scene -- encoding limited interactions in the latent space.

\myparagraph{The Joint Prior.} The prior term in \cref{eq:jcvae}, $p_{\theta}( \mathbf{z}_i | \mathbf{Z}_{< i}, \mathbf{Y}_{< i}, \mathbf{X})$, encodes the effect of interactions on the latent space of agent $i$ through the dependence on $\mathbf{Z}_{< i}, \mathbf{X} $.
In practice, we find that a simpler joint prior,
\begin{align}\label{eq:jcvae_prior}
    \begin{split}
        p_{\theta}(\mathbf{Z} | \mathbf{X}) = p_{\theta}(\mathbf{z}_{1}  | \mathbf{X} ) \cdots p_{\theta}(\mathbf{z}_{n}  | \mathbf{Z}_{<n}, \mathbf{X})
    \end{split}
\end{align}
is sufficient for rich latent spaces that captures interactions.
We parameterize the prior as a conditional normal distribution, where the mean and variance depends on $\{\mathbf{Z}_{< i}, \mathbf{X}\}$. 

\myparagraph{The ELBO.} As the standard log evidence lower bound (ELBO) for cVAEs, proposed in \cite{KingmaW13,SohnLY15}, experiences several issues \eg posterior collapse, we employ the ELBO formulation of $\beta$-VAE \cite{HigginsMPBGBML17} to improve the representational capacity of the latent space and more accurately capture the effect of interactions in dense urban environments.
With the formulation of the factorized variational distribution $q_{\phi}(\mathbf{Z} | \mathbf{X}, \mathbf{Y})$ in \cref{eq:jcvae_posterior} and the joint prior distribution in \cref{eq:jcvae_prior}, the ELBO is given by (details in the appendix),
\begin{align}\label{eq:jcvae_elbo}\raisetag{1.37cm}
\begin{split}
    \log(p_{\theta}( \mathbf{Y} &| \mathbf{X} )) \geq \sum\limits_{i} \mathbb{E}_{q_{\phi}} \log(p_{\theta} \big(\mathbf{y}_{i} | \mathbf{Z}_{\leq i}, \mathbf{Y}_{< i},  \mathbf{X} )\big)  \\
    &  - \beta \sum\limits_{i} D_{\text{KL}}\big(q_{\phi}(\mathbf{z}_{i} | \mathbf{Z}_{< i},  \mathbf{X}, \mathbf{Y} ) || p_{\theta}(\mathbf{z}_{i}  | \mathbf{Z}_{< i}, \mathbf{X} )\big).
\end{split}
\end{align}
Additionally, we find it beneficial to model the observation noise $\sigma^2$ in the posterior distribution over the trajectories $p_{\theta} \big(\mathbf{y}_{i} | \mathbf{Z}_{\leq i}, \mathbf{Y}_{< i},  \mathbf{X} )$ as recommended in \cite{LucasTG019}.
During training, we  alternately optimize both the posterior and the prior distributions such that the ELBO is maximized \cite{abs-1908-09008,TomczakW18}. 

We find in practice that is it sufficient to condition the decoder $p_{\theta} (\mathbf{y}_{i} | \mathbf{Z}_{\leq i}, \mathbf{Y}_{< i}, \mathbf{X} )$ only on $\{ \mathbf{Z}_{\leq i}, \mathbf{X} \}$. The rich latent space of our Joint-$\beta$-cVAE already models the effect of interactions, making it unnecessary to additionally condition on $\mathbf{Y}_{< i}$ for good performance. 

\begin{table*}[t!]
  \centering
  \footnotesize
  \begin{tabularx}{\textwidth}{@{\extracolsep{3pt}}Xcccccccc@{}}
	\toprule
	& \multicolumn{2}{c}{\textbf{Interactions}} & \multicolumn{3}{c}{\textbf{Best of $N\!=\!20$} $\downarrow$} & \multicolumn{3}{c}{\textbf{KDE NLL} $\downarrow$} \\
	\cline{2-3} \cline{4-6}  \cline{7-9} 
	\textbf{Method} &  \textbf{P-P} &  \textbf{P-V} & \textbf{$t+1$ sec} & \textbf{$t+2$ sec} & \textbf{$t+3$ sec} & \textbf{$t+1$ sec} & \textbf{$t+2$ sec} & \textbf{$t+3$ sec} \Tstrut\Bstrut\\
	\midrule
	Social-GAN \cite{GuptaJFSA18} & \checkmark & -- & 0.04 & 0.11 & 0.21 & -2.78 & -1.40 & -0.46 \\
	Social-GAN \cite{GuptaJFSA18} & \checkmark & \checkmark & 0.04 & 0.11 & 0.21 & -2.80 & -1.41 & -0.48\\
	\midrule
	Sophie \cite{SadeghianKSHRS19} & \checkmark & -- & 0.04 & 0.11 & 0.21 & -2.59 & -1.26 & -0.41 \\
	Sophie \cite{SadeghianKSHRS19} & \checkmark & \checkmark & 0.04 & 0.11 & 0.21 & -2.63 & -1.27 & -0.42  \\
	\midrule
	Trajectron++ \cite{TrajTim20}  & \checkmark & -- & \textbf{0.01} & 0.08 & 0.15 & -5.55 & -3.87 & -2.69\\
	Trajectron++ \cite{TrajTim20}  & \checkmark & \checkmark  & \textbf{0.01} & 0.08 & 0.15 & -5.58 & -3.96 & -2.77\\
	\midrule
	cVAE & -- & -- & 0.05 & 0.12 & 0.23 & -2.51 & -1.20 & -0.21 \\
	$\beta$-cVAE \cite{HigginsMPBGBML17}  & -- & -- & \textbf{0.01} & 0.08 & 0.17 & -6.90 & -4.10 & -2.41 \\
	Joint-$\beta$-cVAE (Ours)  & \checkmark & -- & \textbf{0.01} & \textbf{0.06} & 0.13 & -7.50 & -4.53 & -2.95\\
	Joint-$\beta$-cVAE (Ours)  & \checkmark & \checkmark  & \textbf{0.01} & \textbf{0.06} & \textbf{0.13} & \textbf{-7.55} & \textbf{-4.59} & \textbf{-2.98} \\
	\bottomrule
  \end{tabularx}
  \caption{Evaluation on nuScenes. P-P and P-V: whether pedestrian - pedestrian or pedestrian - ego-vehicle interactions are modeled.}
  \vspace{-0.2cm}
  \label{tab:nuScenes_eval}
\end{table*}

\begin{table*}[t]
  \centering
  \footnotesize
  \begin{tabularx}{\textwidth}{@{\extracolsep{3pt}}Xcccccccc@{}}
	\toprule
	& \multicolumn{2}{c}{\textbf{Interactions}} & \multicolumn{3}{c}{\textbf{Best of $N\!=\!20$} $\downarrow$} & \multicolumn{3}{c}{\textbf{KDE NLL} $\downarrow$} \\
	\cline{2-3} \cline{4-6}  \cline{7-9} 
	\textbf{Method} &  \textbf{P-P} &  \textbf{P-V} & \textbf{$t+1$ sec} & \textbf{$t+2$ sec} & \textbf{$t+3$ sec} & \textbf{$t+1$ sec} & \textbf{$t+2$ sec} & \textbf{$t+3$ sec} \Tstrut\Bstrut\\
	\midrule
	Social-GAN \cite{GuptaJFSA18} & \checkmark & --  & 0.14 & 0.36 & 0.65 & -0.38 & 0.74 & 1.55\\
	Social-GAN \cite{GuptaJFSA18} & \checkmark & \checkmark & 0.14 & 0.35 & 0.64 & -0.41 & 0.73 & 1.52 \\
	\midrule
	Sophie \cite{SadeghianKSHRS19} & \checkmark & -- &  0.11 & 0.30 & 0.58  & -1.53 & -0.22 & 0.53 \\
	Sophie \cite{SadeghianKSHRS19} & \checkmark & \checkmark & 0.11 & 0.29 & 0.56 & -1.71 & -0.31 & 0.40  \\
	\midrule
	Trajectron++ \cite{TrajTim20}  & \checkmark & -- & \textbf{0.09} & 0.29 & 0.56 & -2.75 & -0.91 & 0.23  \\
	Trajectron++ \cite{TrajTim20}  & \checkmark & \checkmark  & \textbf{0.09} & 0.28 & 0.54 & -2.81 & -1.00 & 0.15 \\
	\midrule
	cVAE & -- & -- & 0.12 & 0.32 & 0.60 & -1.40 & -0.08 & 0.78 \\
	$\beta$-cVAE \cite{HigginsMPBGBML17}  & -- & -- & 0.10 & 0.30 & 0.56 & -2.61 & -0.78 & 0.31  \\
	Joint-$\beta$-cVAE (Ours)  & \checkmark & -- & \textbf{0.09} & 0.29 & 0.53 & -3.69 & -1.29 & 0.02 \\
	Joint-$\beta$-cVAE (Ours)  & \checkmark & \checkmark  & \textbf{0.09} & \textbf{0.27} & \textbf{0.51} & \textbf{-3.75} & \textbf{-1.38} & \textbf{-0.05} \\
	\midrule
	Joint-$\beta$-cVAE + \{Camera, Lidar\} (Ours) & \checkmark & \checkmark  & \textit{0.09} & \emph{0.27} & \emph{0.50} & \emph{-3.78} & \emph{-1.41} & \emph{-0.13} \\
	\bottomrule
  \end{tabularx}
  \caption{Evaluation on Euro-PVI. P-P and P-V: whether pedestrian - pedestrian or pedestrian - ego-vehicle interactions are modeled.}
  \vspace{-0.2cm}
  \label{tab:toyota_eval}
\end{table*}

\begin{figure*}[t]
\centering
\begin{subfigure}[t]{.3\textwidth}
  \centering
  \includegraphics[height=3.25cm]{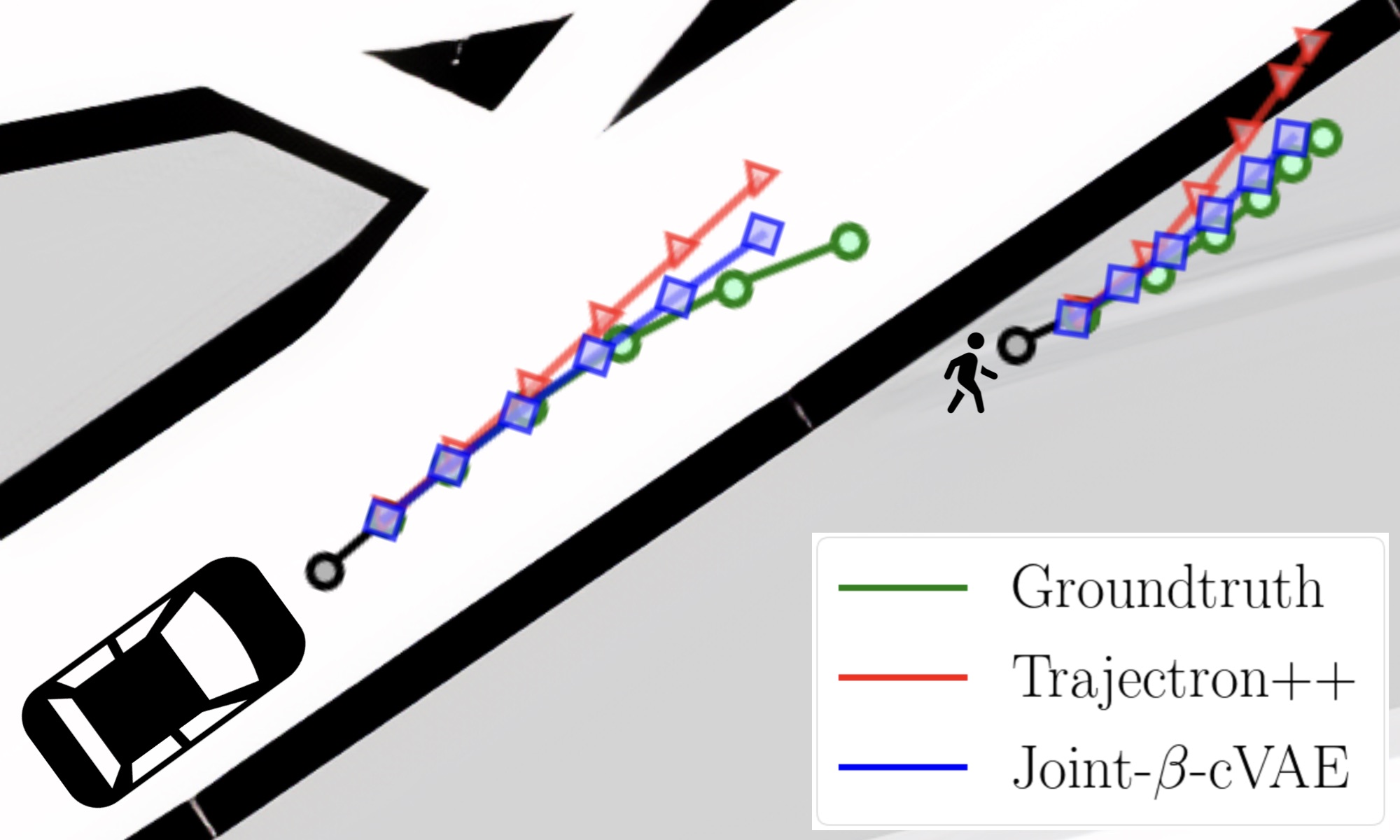}
\end{subfigure}%
\hfill
\begin{subfigure}[t]{.3\textwidth}
  \centering
  \includegraphics[height=3.25cm]{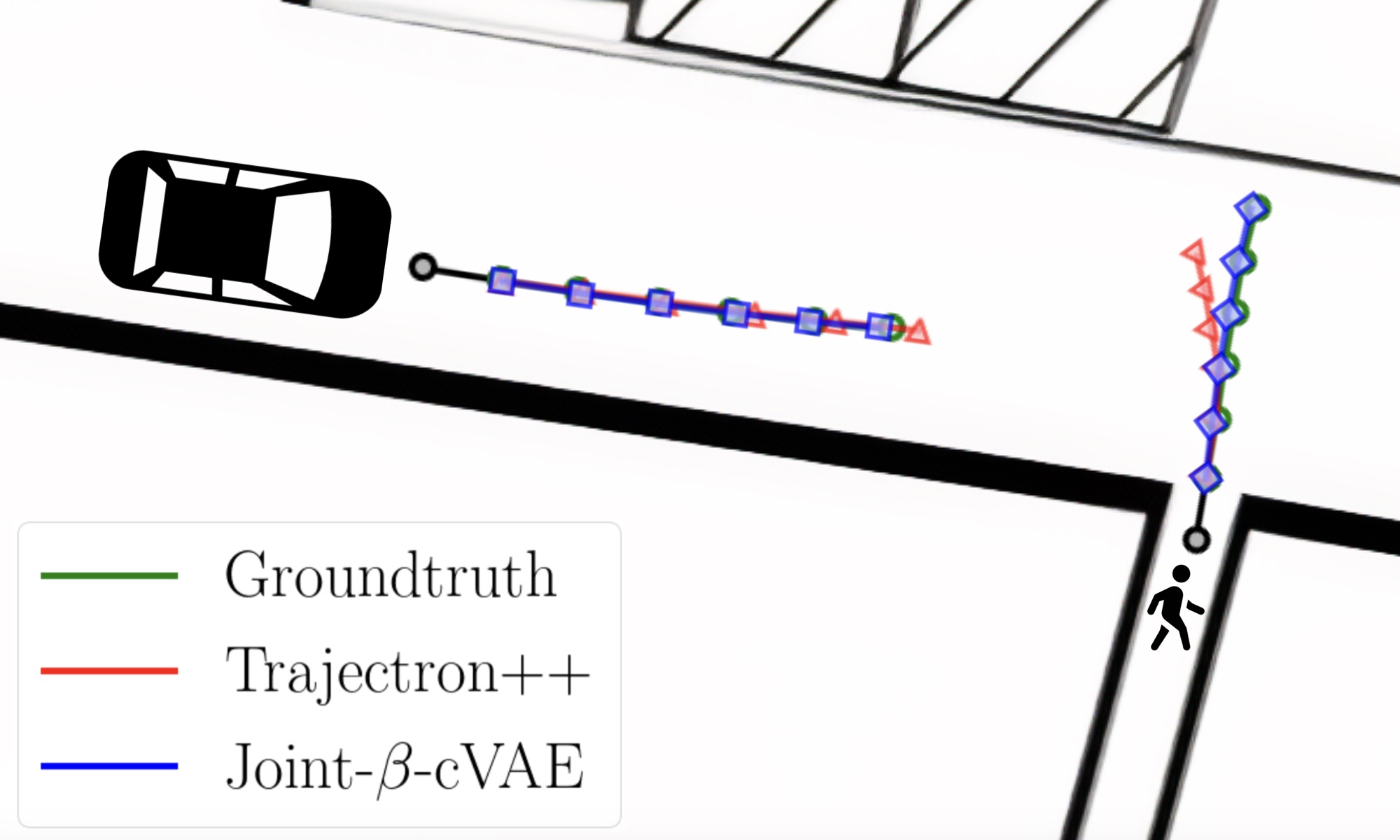}
\end{subfigure}
\hfill
\begin{subfigure}[t]{.3\textwidth}
  \centering
  \includegraphics[height=3.25cm]{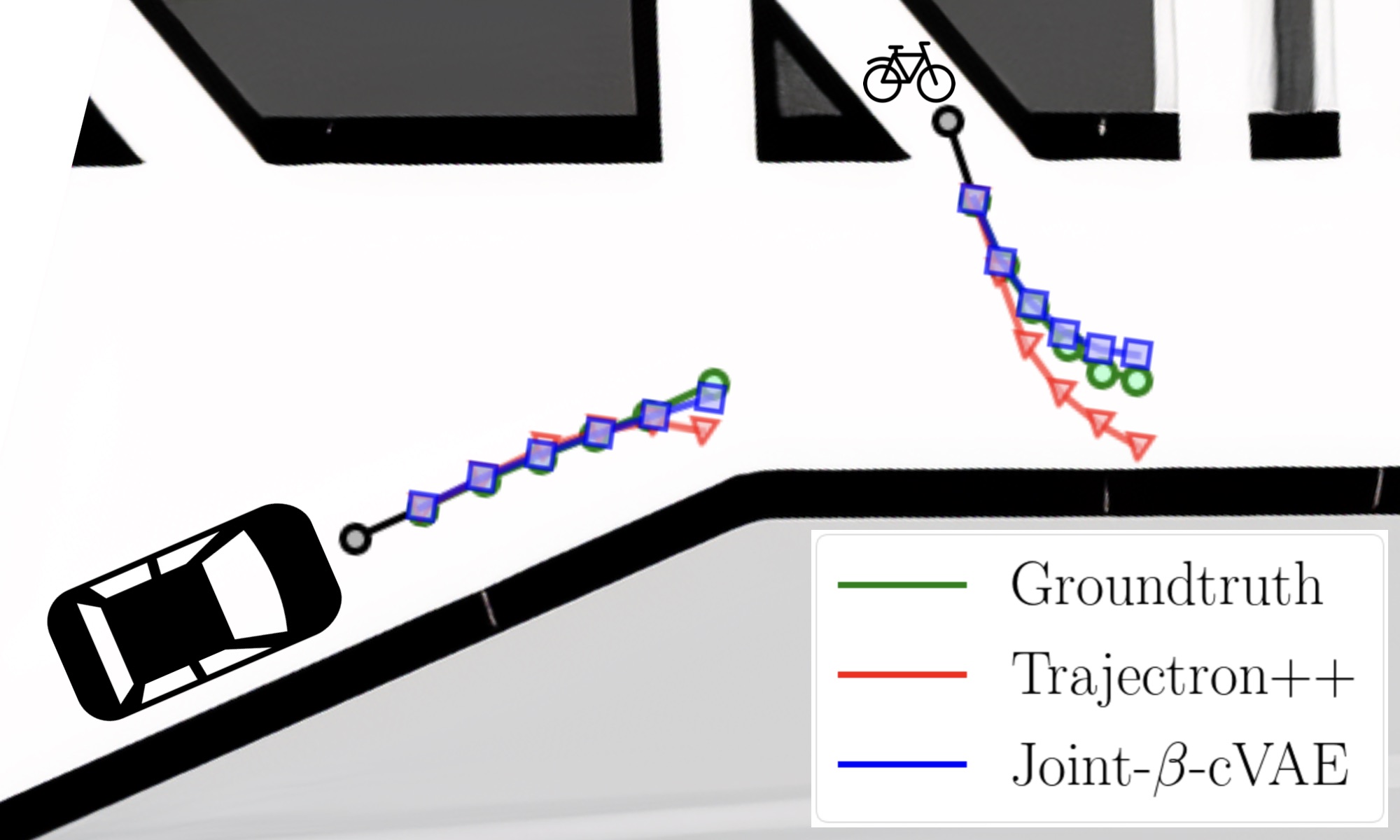}
\end{subfigure}%
\caption{Qualitative examples on Euro-PVI. We compare the Best of $N\!=\!20$ samples for Trajectron++ (red) and our Joint-$\beta$-cVAE (blue).}
\label{fig:q_examp}
\end{figure*}

\section{Experiments}
In this section, we \begin{enumerate*}
    \item Demonstrate the effectiveness of our Joint-$\beta$-cVAE method.
    \item Provide additional experimental evidence to better highlight the differences in the density of interactions between the ego-vehicle and pedestrians(bicyclists) Euro-PVI and current datasets.
\end{enumerate*}
In order to address the above points, in addition to Euro-PVI, we evaluate on nuScenes \cite{CaesarBLVLXKPBB20}. We choose nuScenes \cite{CaesarBLVLXKPBB20} as it is significantly larger compared to datasets like ApolloScapes \cite{MaZZYWM19} and more diverse in comparison to Lyft L5 \cite{abs-2006-14480}, while possessing similar proximity/acceleration statistics (\cref{fig:closestapproach}). 
We first evaluate our approach on nuScenes dataset followed by the evaluation on our proposed Euro-PVI dataset. 

\myparagraph{Evaluation Metrics.} Following \cite{TrajTim20}, we report, \begin{enumerate*}
    \item Best of $N$ (FDE): The final (euclidean) displacement error in meters using the best of $N\!=\!20$ samples \cite{GuptaJFSA18,SadeghianKSHRS19,TrajTim20}.
    \item KDE NLL: The (mean) negative log-likelihood of the groundtruth trajectory under the predicted distribution estimated using a Gaussian kernel \cite{IvanovicP19,ThiedeB19}, computed using the code provided by \cite{TrajTim20}.
\end{enumerate*} Both these metrics aim to measure the match of the predicted trajectory distribution to the groundtruth distribution \cite{abs-1908-09008,GuptaJFSA18,LeeCVCTC17}. We provide results with the average (euclidean) displacement error in the appendix.

\myparagraph{Implementation Details.} We provide additional implementation details \eg number of layers, hyper-parameters and code in the appendix.

\subsection{nuScenes}
Following \cite{TrajTim20}, we split the training set into the training and validation splits. The original validation split is used as test set. 
We provide 1 - (\emph{upto}) 5 secs of observation (historical context) and predict 3 seconds ahead \cite{TrajTim20}. 
We compare our Joint-$\beta$-cVAE approach to the following state of the art models: Social-GAN \cite{GuptaJFSA18}, Sophie \cite{SadeghianKSHRS19} and Trajectron++ \cite{TrajTim20}.
Additionally, in order to measure the density and influence of interactions between the ego-vehicle and pedestrians (bicyclists) on the trajectories in nuScenes (in comparison to Euro-PVI), we also evaluate the above methods without modeling ego-vehicle - pedestrians (bicyclists) interactions. 
Any significant difference in performance of these models would indicate the presence of dense ego-vehicle - pedestrian (bicyclist) interactions. 

To illustrate the effectiveness of our Joint-$\beta$-cVAE, we also include two  ablations of our Joint-$\beta$-cVAE model, \begin{enumerate*} 
\item A simple cVAE model and, 
\item A $\beta$-cVAE model,
\end{enumerate*} (neither of which can model interactions). These ablations are designed to show the effectiveness of our Joint-$\beta$-cVAE model in capturing interactions in the latent space.

We report results using both the Best of $N$ and KDE NLL metrics in \cref{tab:nuScenes_eval}. The P-P and P-V columns in \cref{tab:nuScenes_eval} indicate whether the pedestrian (bicyclist) - pedestrian (bicyclist) and pedestrian( bicyclist) - ego-vehicle interactions are modeled -- using social pooling in case of Social-GAN \cite{GuptaJFSA18}, the attention mechanism in case of Sophie \cite{SadeghianKSHRS19}, the scene graph in case of Trajectron++ \cite{TrajTim20} and with a shared latent space in case of our Joint-$\beta$-cVAE model. Trajectron++ outperforms both the conditional GAN based Social-GAN and Sophie models -- partly due to the better modeling of interaction with the scene graph compared to Social-GAN and Sophie. Also, note that Trajectron++ is built using a cVAE backbone. Thus, the performance advantage of Trajectron++ also illustrates the effectiveness of cVAE based models in capturing the distribution of future trajectories. We see that our Joint-$\beta$-cVAE outperforms Trajectron++. Additionally, our Joint-$\beta$-cVAE outperforms  both the simple cVAE and $\beta$-cVAE ablations, illustrating that our Joint-$\beta$-cVAE model can effectively model interactions in the latent space. The performance advantage of our Joint-$\beta$-cVAE model over Trajectron++ shows the advantage of a joint latent space that can model the effect of interactions, in comparison to independent latent spaces which model the effect of interactions only as an additional condition to the decoder. 
Finally, across all models, we see that models which additionally encode pedestrian (bicyclist) - ego-vehicle interactions in nuScenes do not show a significant improvement in performance. This further lends support to the fact that pedestrian (bicyclist) - ego-vehicle interactions are sparse in the nuScenes dataset.

\subsection{Euro-PVI}
We now evaluate the different models for trajectory prediction on our novel Euro-PVI dataset. We use 792 sequences for training, 100 sequences for validation and 185 sequences for testing. The train/val/test splits do not share pedestrian (bicyclist) instances.
As on nuScenes, we compare our Joint-$\beta$-cVAE model with the Social-GAN \cite{GuptaJFSA18}, Sophie \cite{SadeghianKSHRS19}, Trajectron++ \cite{TrajTim20} models. 
We also include the cVAE and $\beta$-cVAE ablations (which cannot model interactions), to establish whether our Joint-$\beta$-cVAE approach can model interactions in the latent space.
We follow a similar evaluation protocol as in nuScenes, where we predict trajectories up to 3 seconds into the future. However, we provide a shorter observation of 0.5 seconds as quick reactions are essential in dense traffic scenarios.
 
We report the results in \cref{tab:toyota_eval}.  
As on nuScenes, we observe that our Joint-$\beta$-cVAE approach outperforms the competing methods. 
The performance advantage over Trajectron++ again illustrates the advantage of the joint latent space over all agents in the scene versus an independent latent space which cannot model the effect of interactions.
Our Joint-$\beta$-cVAE  outperforms the cVAE and $\beta$-cVAE baselines, which illustrates that our Joint-$\beta$-cVAE model can model the effect of interactions successfully in the latent space. Additionally, the performance gain on Euro-PVI  (0.03m, Best of $N\!=\!20$) 
of our Joint-$\beta$-cVAE model over Trajectron++ is larger than in nuScenes. This shows that our Joint-$\beta$-cVAE model can better capture the complex distribution of pedestrians (bicyclists) trajectories in dense urban environments under the effect of interactions.  We further show that performance can be improved by conditioning our Joint-$\beta$-cVAE model on visual features from the camera and lidar (also see appendix). We provide qualitative examples comparing our Joint-$\beta$-cVAE model to Trajectron++ in \cref{fig:q_examp}. We see in \cref{fig:q_examp} (left) that Trajectron++ incorrectly predicts that the pedestrian will step onto the road, while our Joint-$\beta$-cVAE correctly predicts that due to the oncoming ego-vehicle the pedestrian avoids stepping onto the road. Similarly, in \cref{fig:q_examp} (middle) our Joint-$\beta$-cVAE correctly predicts that the pedestrian quickly crosses the street due to the oncoming ego-vehicle and in \cref{fig:q_examp} (right) the bicyclist merges in front of the ego-vehicle which slows down.

\begin{figure}
\centering
\begin{subfigure}[t]{0.5\linewidth}
  \centering
  \includegraphics[width=\linewidth]{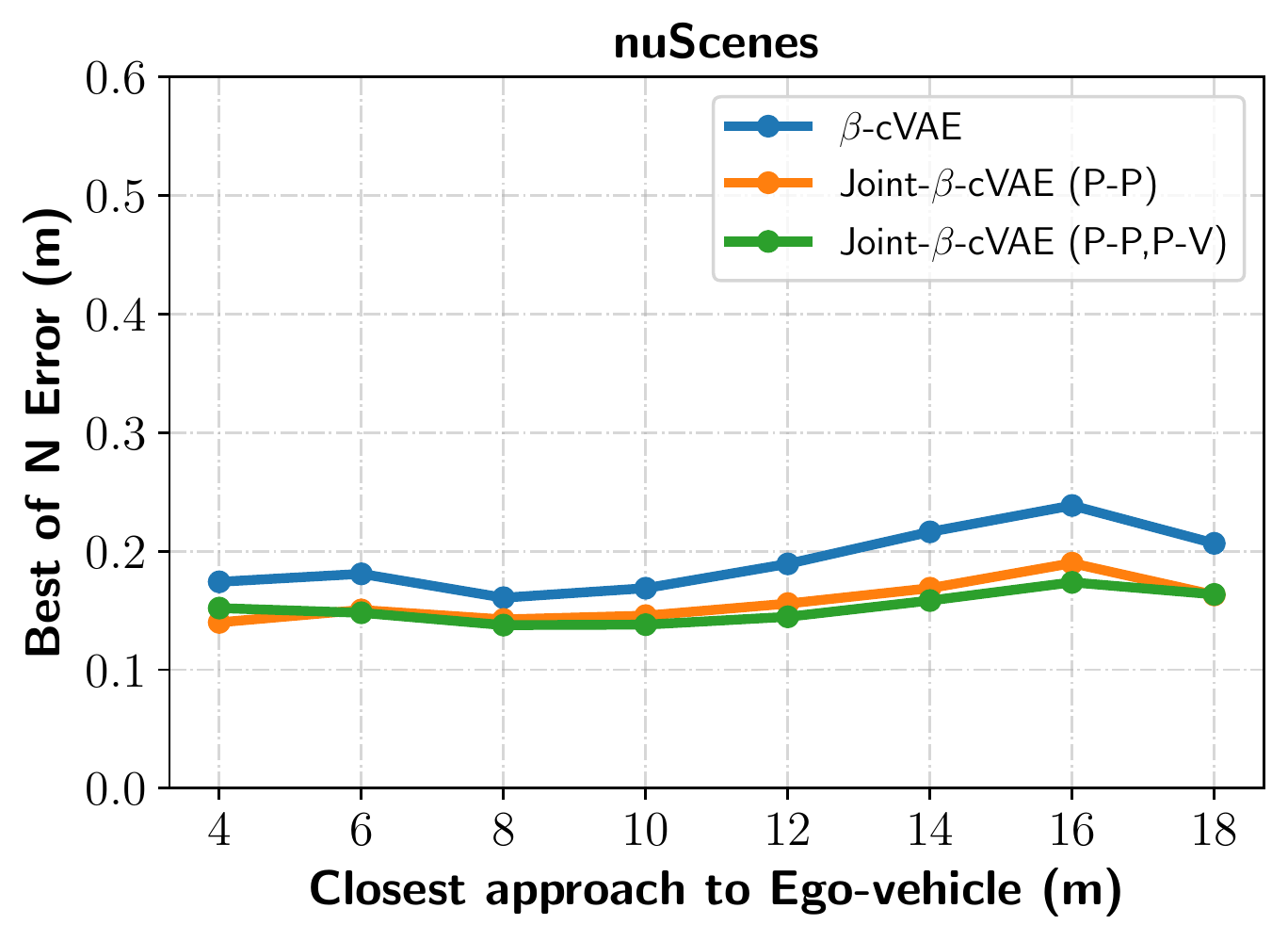}
\end{subfigure}%
\hfill
\begin{subfigure}[t]{0.5\linewidth}
  \centering
  \includegraphics[width=\linewidth]{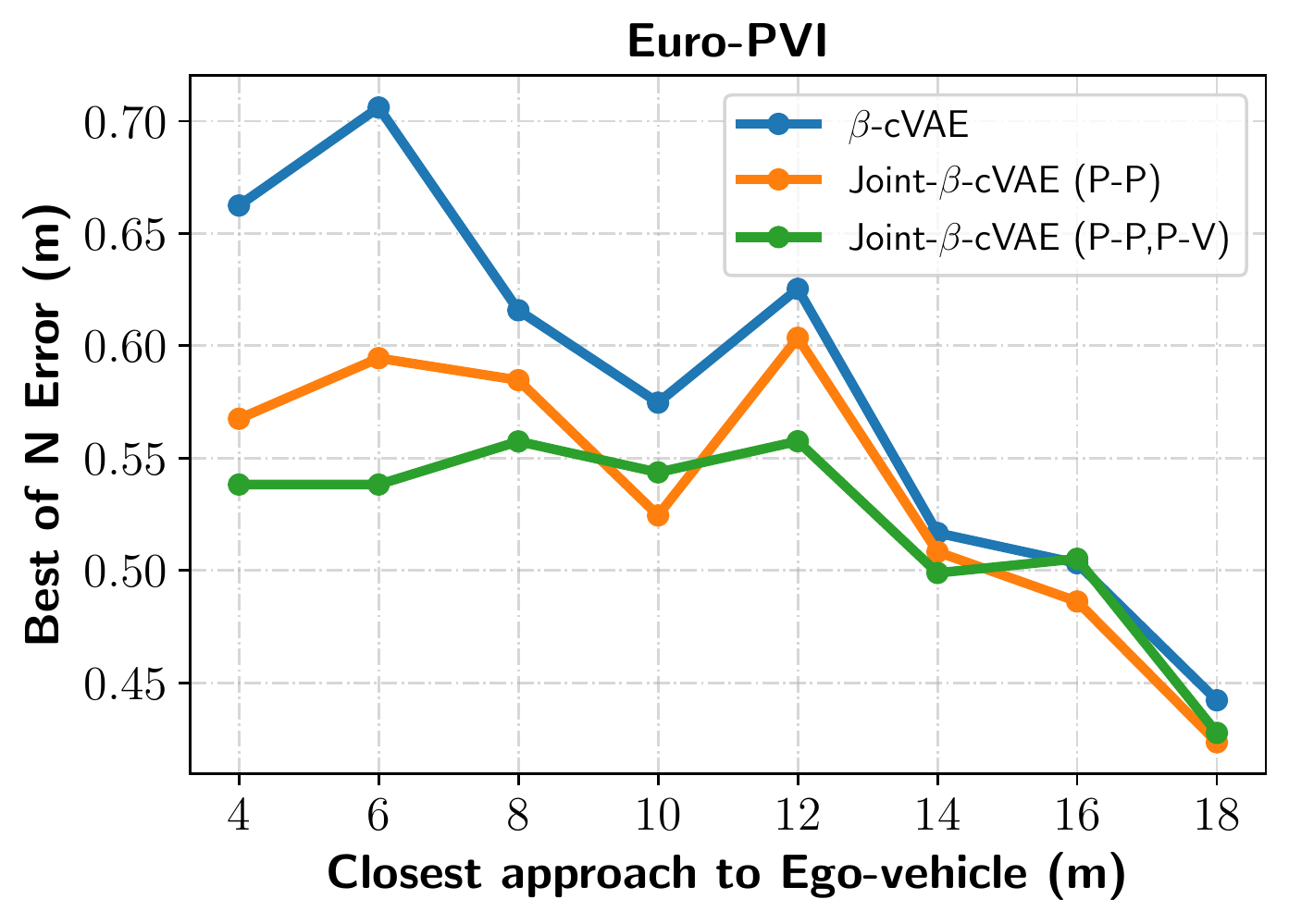}
\end{subfigure}
\caption{Error sorted by closest approach (proximity) to ego-vehicle. Higher error in close proximity to the ego-vehicle suggest dense interactions.}
\label{fig:error_vs_distance}
\end{figure}

Finally, across all methods, we see the gain in performance (using both the Best of $N$ and KDE NLL metrics) across methods when pedestrian (bicyclist) - ego-vehicle (P-V) interactions are modeled in \cref{tab:toyota_eval} is larger than in nuScenes. This provides further evidence of dense pedestrian(bicyclist) - ego-vehicle interactions in Euro-PVI compared to the sparse interactions in nuScenes. Additionally, in \cref{fig:error_vs_distance} we plot the Best of $N$ error of our Joint-$\beta$-cVAE model along with the $\beta$-cVAE ablation versus the distance of the trajectory from the ego-vehicle for both nuScenes and Euro-PVI. We see that in case of nuScenes, the error does not depend strongly on distance. This mirrors the results in \cref{fig:closestapproach} (right) which again suggests that the interactions between pedestrians and the ego-vehicle are sparse. In contrast, in case of our Euro-PVI dataset, the error is largest when the distance between the pedestrian (bicyclist) and the ego-vehicle is smallest \ie at close encounters. This again suggests the presence of dense pedestrian (bicyclist) - ego-vehicle interactions in Euro-PVI.

\begin{table}[h]
  \centering
   \footnotesize
  \begin{tabularx}{\linewidth}{@{}Xccc@{}}
	\toprule
	& \multicolumn{3}{c}{\textbf{Best of $N\!=\!20$} $\downarrow$}\\
	\cline{2-4}
	\textbf{Method} &  \textbf{$t+1$ sec} & \textbf{$t+2$ sec} & \textbf{$t+3$ sec} \Tstrut\Bstrut\\
	\midrule
	Trajectron++ \cite{TrajTim20}  & 0.10 & 0.35 & 0.63 \\
	Joint-$\beta$-cVAE (Ours)  & 0.10 & 0.33 & 0.61\\
	\bottomrule
  \end{tabularx}
  \caption{Transferring models trained on nuScenes to Euro-PVI.}
  \label{tab:toyota_to_blip_eval}
\end{table}

\myparagraph{Transferring Models from nuScenes.} Finally, we experiment with transferring the best performing models on nuScenes \ie Trajectron++ and our Joint-$\beta$-cVAE from nuScenes (with both P-P,P-V interactions) to Euro-PVI in \cref{tab:toyota_to_blip_eval}. We observe a considerable drop in performance in the Best of $N$ error in comparison to the performance of the models when they are \emph{both} trained and evaluated on Euro-PVI (\cref{tab:toyota_eval}). This provides additional evidence that the distribution of trajectories and interaction patterns in Euro-PVI is significantly different compared to nuScenes. We provide additional results in the appendix.

\section{Conclusion and Outlook}
We presented Euro-PVI, a new dataset with dense scenarios of vehicle-pedestrian (bicyclist) interaction and their trajectories to advance the task of future trajectory prediction which is crucial to the development of self-driving vehicles. 
We investigated the effect of interactions in urban environments on current state-of-the-art methods for existing nuScenes dataset which show a  notable performance gap on our Euro-PVI dataset.
To address this challenge of modeling complex interactions, we propose a Joint-$\beta$-cVAE approach. 
We demonstrate state of the art results both on nuScenes and on Euro-PVI. 
The performance advantage of our Joint-$\beta$-cVAE on Euro-PVI highlights the effectiveness of our approach in dense urban scenarios.
The key to our success is a shared latent space between the interacting agents -- which encodes the effect of intersections -- in comparison to prior work which employ conditionally independent latent spaces.
We believe that the Euro-PVI dataset along with Joint-$\beta$-CVAE approach provide a new important dimension to the task of future trajectory prediction with dense ego-vehicle - pedestrian (bicyclist) interactions. 

{\small
\bibliographystyle{ieee_fullname}
\bibliography{egbib}
}

\clearpage
\appendix
\appendixpage
\addappheadtotoc

\section{Introduction}
In the appendix, we include \begin{enumerate*}
    \item Further details about the Euro-PVI dataset (in \cref{sec:europvi_details}).
    \item Further details of our Joint-$\beta$-cVAE and additional evaluation (in \cref{sec:jcvae_details}).
\end{enumerate*}

In particular, in \cref{sec:europvi_details} we provide further details of the sensor setup, adding to the information in Section 3 of the main paper and also clarify the collection process of the Euro-PVI dataset.
To further highlight the diversity of ego-vehicle - pedestrian (bicyclist) interactions in dense urban scenes, we include additional qualitative examples of the interactions in the Euro-PVI dataset in \cref{fig:example_interactions}.  

In \cref{sec:jcvae_details}, we elaborate the deviation of the ELBO for our Joint-$\beta$-cVAE approach (\cf Eq.~(5) in the main paper) and provide additional architectural details of our Joint-$\beta$-cVAE model (\cf Sec.~5 in the main paper).
Additionally, we also provide details of conditioning our Joint-$\beta$-cVAE model on visual features (\cf Table 3 in the main paper), evaluation using ADE and FDE metrics with sample sizes of $N\!=\!\{3,20\}$ (\cref{tab:toyota_eval_addmat_n3,tab:toyota_eval_addmat_n20}), evaluation using the KDE NLL metric of models transferred from nuScenes \cite{CaesarBLVLXKPBB20} to Euro-PVI (\cref{tab:toyota_to_blip_eval}) and evaluation of  models trained both on nuScenes and Euro-PVI on nuScenes (\cref{tab:toyota_to_nuscenes_eval}).
We also provide qualitative examples to further highlight the effectiveness of Joint-$\beta$-cVAE model in capturing ego-vehicle - pedestrian (bicyclist) interactions in \cref{fig:q_examp}.

\section{Further Details about the Euro-PVI dataset}\label{sec:europvi_details}

\myparagraph{Further Details of Sensor Setup for Euro-PVI.}
The Euro-PVI dataset was recorded from a vehicle equipped with a lidar (Velodyne HDL-64E) mounted over the roof, which captures point clouds with a frequency of 10Hz. The vehicle also includes front-facing camera(s) installed behind the windshield which have a similar point of view as the driver. The cameras capture images at a resolution of $1280\! \times \! 806$, and a frequency of 10Hz. Images are calibrated to remove distortion. This setup is adequate for capturing ego-vehicle - pedestrian (bicyclist) interactions as most interactions happen in front of the vehicle. Further, all sensors are registered to a common frame coordinate system inside the vehicle. Each data point has a corresponding position/pose from an on-board high performance GPS/IMU, and all are timestamped and synchronized. Such a setup also allows for the use of mapping services \eg OpenStreetMap.

\myparagraph{Examples of Interactions in the Euro-PVI dataset.}
We provide additional examples of interactions in the Euro-PVI dataset in \cref{fig:example_interactions} with five example interactions between the ego-vehicle and pedestrians (bicyclists) to highlight the diversity of interactions. Analogous to the Fig.~2 in the main paper, we also include the $L_2$ norms of the velocity and acceleration to illustrate the effect of interactions on the pedestrian (bicyclist) trajectories -- which again highlights the need to model such ego-vehicle - pedestrian (bicyclist) interactions for accurate pedestrian (bicyclist) trajectory prediction.

\section{Further Details of our Joint-$\beta$-CVAE and Additional Evaluation}\label{sec:jcvae_details}
We now provide additional details of our Joint-$\beta$-cVAE approach. We first detail the derivation of the ELBO (Eq.~(5) in the main paper). Following this, we provide the details of the network architecture and the hyperparameters used in the  Joint-$\beta$-cVAE model.

\myparagraph{Details of the ELBO.} From Eq.~(2) in the main paper, the joint probability of the future trajectories of the agents in the scene can be expressed as,
\begin{align}\label{eq:jcvae_r}\raisetag{+1.4cm}
    \begin{split}
        p_{\theta}&(\mathbf{Y} | \mathbf{X} ) \\ &= \int \prod\limits_i^n p_{\theta}(\mathbf{y}_{i} | \mathbf{Z}_{\leq i}, \mathbf{Y}_{< i}, \mathbf{X} ) p_{\theta}( \mathbf{z}_i | \mathbf{Z}_{< i}, \mathbf{Y}_{< i}, \mathbf{X})  \diff \mathbf{Z}.
    \end{split}
\end{align}
Using the joint posterior as defined in Eq.\ (3) in the main paper, the joint probability above can be expressed as,
\begin{align}\label{eq:jcvae_pos}
    \begin{split}
        p_{\theta}&(\mathbf{Y} | \mathbf{X} ) \\ &= \int \prod\limits_i^n p_{\theta}(\mathbf{y}_{i} | \mathbf{Z}_{\leq i}, \mathbf{Y}_{< i}, \mathbf{X} ) \frac{p_{\theta}( \mathbf{z}_i | \mathbf{Z}_{< i}, \mathbf{Y}_{< i}, \mathbf{X})}{q_{\phi}(\mathbf{z}_{i}  | \mathbf{Z}_{< i},  \mathbf{X}, \mathbf{Y} )} \\
        &q_{\phi}(\mathbf{z}_{i}  | \mathbf{Z}_{< i},  \mathbf{X}, \mathbf{Y} )  \diff \mathbf{Z}.
    \end{split}
\end{align}
Therefore, the log-likelihood of the joint distribution is,
\begin{align}\label{eq:jcvae_log}\raisetag{+1.7cm}
    \begin{split}
        &\log(p_{\theta}(\mathbf{Y} | \mathbf{X} )) \\ &= \log\Bigg( \prod\limits_i^n \mathbb{E}_{q_{\phi}} \Big(  p_{\theta}(\mathbf{y}_{i} | \mathbf{Z}_{\leq i}, \mathbf{Y}_{< i}, \mathbf{X} ) \frac{p_{\theta}( \mathbf{z}_i | \mathbf{Z}_{< i}, \mathbf{Y}_{< i}, \mathbf{X})}{q_{\phi}(\mathbf{z}_{i}  | \mathbf{Z}_{< i},  \mathbf{X}, \mathbf{Y} )} \Big) \Bigg)\\
        \\ &= \sum\limits_i^n \log\Bigg(  \mathbb{E}_{q_{\phi}} \Big(  p_{\theta}(\mathbf{y}_{i} | \mathbf{Z}_{\leq i}, \mathbf{Y}_{< i}, \mathbf{X} ) \frac{p_{\theta}( \mathbf{z}_i | \mathbf{Z}_{< i}, \mathbf{Y}_{< i}, \mathbf{X})}{q_{\phi}(\mathbf{z}_{i}  | \mathbf{Z}_{< i},  \mathbf{X}, \mathbf{Y} )} \Big) \Bigg).\\
    \end{split}
\end{align}

Now, using Jensen's inequality  and introducing the $\beta$ term to weigh the KL-divergence term ($D_{\text{KL}}$) as in \cite{HigginsMPBGBML17} gives us the ELBO in Eq.\ (5) of the main paper,
\begin{align}\notag
\begin{split}
    \log(p_{\theta}( \mathbf{Y} &| \mathbf{X} )) \geq \sum\limits_{i} \mathbb{E}_{q_{\phi}} \log(p_{\theta} \big(\mathbf{y}_{i} | \mathbf{Z}_{\leq i}, \mathbf{Y}_{< i},  \mathbf{X} )\big)  \\
    &  - \beta \sum\limits_{i} D_{\text{KL}}\big(q_{\phi}(\mathbf{z}_{i} | \mathbf{Z}_{< i},  \mathbf{X}, \mathbf{Y} ) || p_{\theta}(\mathbf{z}_{i}  | \mathbf{Z}_{< i}, \mathbf{X} )\big).
\end{split}
\end{align}

\myparagraph{Additional Architectural Details.}
We model the posterior distribution $(q_\phi)$ using LSTMs with 128 hidden neurons. 
The attention over each of the previously sampled $\mathbf{z}_{j} \in \mathbf{Z}_{< i}$ and $\mathbf{x}_j \in \mathbf{X}, \mathbf{y}_j \in \mathbf{Y}$ is modeled using fully connected layers with 64 hidden units. 
Similarly, the prior, $p_\theta$ is modeled using an LSTM with 128 hidden units and the attention over each of the previously sampled $\mathbf{z}_{j} \in \mathbf{Z}_{< i}$ and $\mathbf{x}_j \in \mathbf{X}$ is modeled using fully connected layers with 64 hidden units. 
The decoder is also modeled using an LSTM with 128 hidden units. We use a latent space of 32 dimensions. 
We find that $\beta$ value of $[0.08,0.12]$ works well in practice and helps us learn representative latent spaces. We use the Adam \cite{KingmaB14} optimizer with a learning rate of $3 \times 10^{-3}$ with a exponential decay of $0.9999$.

\begin{table*}[h]
  \centering
  \footnotesize
  \begin{minipage}[t]{\textwidth}
  \begin{tabularx}{\textwidth}{@{}Xcccccccc@{}}
	\toprule
	 & \multicolumn{2}{c}{\textbf{Interactions}} & \multicolumn{3}{c}{\textbf{FDE $N\!=\!20$} $\downarrow$} & \multicolumn{3}{c}{\textbf{ADE $N\!=\!20$} $\downarrow$} \\
	\cline{2-3} \cline{4-6}  \cline{7-9} 
	\textbf{Method} & \textbf{P-P} &  \textbf{P-V} & \textbf{$t+1$ sec} & \textbf{$t+2$ sec} & \textbf{$t+3$ sec} & \textbf{$t+1$ sec} & \textbf{$t+2$ sec} & \textbf{$t+3$ sec} \Tstrut\Bstrut\\
	\midrule
	Trajectron++ \cite{TrajTim20} & \checkmark & \checkmark  & \textbf{0.09} & 0.28 & 0.54 & \textbf{0.05} & 0.13 & 0.24 \\
	Joint-$\beta$-cVAE (Ours) & \checkmark & \checkmark  & \textbf{0.09} & \textbf{0.27} & \textbf{0.51} & \textbf{0.05} & \textbf{0.12} & \textbf{0.23} \\
	\bottomrule
  \end{tabularx}
  \caption{Additional metrics on Euro-PVI with $N\!=\!20$ samples. P-P and P-V: whether pedestrian - pedestrian or pedestrian - ego-vehicle interactions are modeled.}
  \label{tab:toyota_eval_addmat_n20}
\end{minipage}\hfill
\begin{minipage}[t]{\textwidth}
\vspace{0.15cm}
  \centering
  \footnotesize
  \begin{tabularx}{\textwidth}{@{}Xcccccccc@{}}
	\toprule
	 & \multicolumn{2}{c}{\textbf{Interactions}} & \multicolumn{3}{c}{\textbf{FDE $N\!=\!3$} $\downarrow$} & \multicolumn{3}{c}{\textbf{ADE $N\!=\!3$} $\downarrow$} \\
	\cline{2-3} \cline{4-6}  \cline{7-9} 
	\textbf{Method} & \textbf{P-P} &  \textbf{P-V} & \textbf{$t+1$ sec} & \textbf{$t+2$ sec} & \textbf{$t+3$ sec} & \textbf{$t+1$ sec} & \textbf{$t+2$ sec} & \textbf{$t+3$ sec} \Tstrut\Bstrut\\
	\midrule
	Trajectron++ \cite{TrajTim20} & \checkmark & \checkmark  & 0.18 & 0.53 & 1.01 & \textbf{0.09} & 0.22 & 0.42 \\
	Joint-$\beta$-cVAE (Ours) & \checkmark & \checkmark  & \textbf{0.17} & \textbf{0.51} & \textbf{0.99} & \textbf{0.09} & \textbf{0.22} & \textbf{0.41} \\
	\bottomrule
  \end{tabularx}
  \caption{Additional metrics on Euro-PVI with $N\!=\!3$ samples. P-P and P-V: whether pedestrian - pedestrian or pedestrian - ego-vehicle interactions are modeled.}
  \label{tab:toyota_eval_addmat_n3}
\end{minipage}
\begin{minipage}[t]{\textwidth}
\vspace{0.15cm}
  \centering
   \footnotesize
  \begin{tabularx}{\textwidth}{@{}Xcccccccc@{}}
	\toprule
	& \multicolumn{2}{c}{\textbf{Interactions}} & \multicolumn{3}{c}{\textbf{Best of $N\!=\!20$} $\downarrow$} & \multicolumn{3}{c}{\textbf{KDE NLL} $\downarrow$} \\
	\cline{2-3} \cline{4-6}  \cline{7-9} 
	\textbf{Method} &  \textbf{P-P} &  \textbf{P-V} & \textbf{$t+1$ sec} & \textbf{$t+2$ sec} & \textbf{$t+3$ sec} & \textbf{$t+1$ sec} & \textbf{$t+2$ sec} & \textbf{$t+3$ sec} \Tstrut\Bstrut\\
	\midrule
	Trajectron++ \cite{TrajTim20}  & \checkmark & -- & 0.10 & 0.35 & 0.63 & -1.07 & 0.15 & 1.45 \\
	Trajectron++ \cite{TrajTim20}  & \checkmark & \checkmark  & 0.10 & 0.35 & 0.63 & -1.04 & 0.14 & 1.42 \\
	\midrule
	Joint-$\beta$-cVAE (Ours)  & \checkmark & -- & 0.10 & 0.33 & 0.60 & -1.51 & -0.09 & 1.22 \\
	Joint-$\beta$-cVAE (Ours)  & \checkmark & \checkmark  & 0.10 & 0.33 & 0.61 & -1.56 & -0.10 & 1.31\\
	\bottomrule
  \end{tabularx}
  \caption{Transferring models trained on nuScenes to Euro-PVI (see also Table 4 in the main paper).}
  \label{tab:toyota_to_blip_eval}
\end{minipage}
\begin{minipage}[t]{\textwidth}
\vspace{0.15cm}
  \centering
   \footnotesize
  \begin{tabularx}{\textwidth}{@{}Xcccccccc@{}}
	\toprule
	& \multicolumn{2}{c}{\textbf{Training}} & \multicolumn{3}{c}{\textbf{Best of $N\!=\!20$} $\downarrow$} & \multicolumn{3}{c}{\textbf{KDE NLL} $\downarrow$} \\
	\cline{2-3} \cline{4-6}  \cline{7-9} 
	\textbf{Method} &  \textbf{nuScenes} &  \textbf{Euro-PVI} & \textbf{$t+1$ sec} & \textbf{$t+2$ sec} & \textbf{$t+3$ sec} & \textbf{$t+1$ sec} & \textbf{$t+2$ sec} & \textbf{$t+3$ sec} \Tstrut\Bstrut\\
	\midrule
	Joint-$\beta$-cVAE (Ours)  & \checkmark & -- & 0.01 & 0.14 & 0.31 & -0.20 & 1.97 & 3.56\\
	Joint-$\beta$-cVAE (Ours)  & \checkmark & \checkmark & 0.01 & 0.13 & 0.30 & -0.27 & 1.64 & 2.92 \\
	\bottomrule
  \end{tabularx}
  \caption{Evaluating models trained on both on nuScenes and Euro-PVI on nuScenes (only pedestrian - pedestrian interactions are modeled).}
  \label{tab:toyota_to_nuscenes_eval}
  \end{minipage}
\end{table*}

\myparagraph{Details of Conditioning on Visual Features.}
In Table 3 in the main paper, we report results on our Euro-PVI dataset where our Joint-$\beta$-cVAE model is additionally conditioned on visual features. We use both the RGB camera images and lidar point clouds. In detail, we use a $256\times256$ crop of the camera image and a $5 \text{ meter} \times 5 \text{ meter}$ bird eye view rendering of the lidar point cloud both centered at the pedestrian (bicyclist). The latent space of our Joint-$\beta$-cVAE is additionally conditioned on these visual features using a simple VGG-16 \cite{SimonyanZ14a} like neural network. We see that the performance of our Joint-$\beta$-cVAE further improves, because the camera image and bird eye view lidar provides important contextual information \eg physical obstacles in the vicinity of the pedestrian (bicyclist) which can have a significant impact on the trajectory of the pedestrian (bicyclist).

\myparagraph{Additional Metrics on Euro-PVI.} Here, we additionally report for Trajectron++ \cite{TrajTim20} and our Joint-$\beta$-cVAE, the average (euclidean) displacement error (ADE) at $t+\{1,2,3\}$ seconds for $N\!=\!20$ samples in \cref{tab:toyota_eval_addmat_n20} and both the final (euclidean) displacement error (FDE, equivalent to the Best of $N$ error in Table 2,3 of the main paper) and average (euclidean) displacement error (ADE) at $t+\{1,2,3\}$ seconds for $N\!=\!3$ samples in \cref{tab:toyota_eval_addmat_n3}. The results are consistent with the findings in Table 2,3 of the main paper where our Joint-$\beta$-cVAE outperforms Trajectron++ \cite{TrajTim20} for both sample sizes $N\!=\!\{3,20\}$.

\myparagraph{Additional Metrics on Transferring from nuScenes to Euro-PVI.} While in Table 4 (in the main paper), we report only the Best of $N$ metric, in \cref{tab:toyota_to_blip_eval}, we additionally report the KDE NLL metric. Similar to the observations with the Best of $N$ metric, observe a considerable drop in performance as measured by the KDE NLL metric in comparison to the performance of the models when they are both trained and evaluated on Euro-PVI. Again, this provides additional evidence that the distribution of trajectories and interaction patterns in Euro-PVI is significantly different compared to nuScenes.

\myparagraph{Transferring from Euro-PVI to nuScenes.} In Table 4 (in the main paper), we show that models trained (only) on nuScenes \cite{CaesarBLVLXKPBB20} do not perform well when evaluated on Euro-PVI. On the other hand, in \cref{tab:toyota_to_nuscenes_eval}, we show that training both on Euro-PVI and nuScenes can improve performance on nuScenes. In particular, we consider the more challenging setting we provide a shorter observation of 1 second to our Joint-$\beta$-cVAE. The performance on nuScenes improves because Euro-PVI also contains significant pedestrian (bicyclist) - pedestrian (bicyclist) interactions in addition to vehicle - pedestrian (bicyclist) interactions.

\myparagraph{Further Qualitative Examples.}
To further validate our Joint-$\beta$-cVAE approach for modeling complex interactions in dense urban scenarios, we provide additional qualitative examples of predictions on the Euro-PVI dataset in \cref{fig:q_examp}. 
We compare the predictions of our Joint-$\beta$-cVAE approach to that of Trajectron++ \cite{TrajTim20} using the Best of $N\!=\!20$ samples. 
In \cref{fig:q_examp} (top left) we see that our Joint-$\beta$-cVAE model correctly predicts that the pedestrian stays on the sidewalk and waits for the ego-vehicle to pass. Similarly, in \cref{fig:q_examp} (top right) our Joint-$\beta$-cVAE model correctly predicts that the pedestrian crosses the street in front of the ego-vehicle.
In \cref{fig:q_examp} (middle left) our Joint-$\beta$-cVAE model correctly predicts that the bicyclist yields to the on-coming ego-vehicle.
In \cref{fig:q_examp} (middle right) our Joint-$\beta$-cVAE model correctly predicts that the bicyclist continues straight ahead and does not attempt to cross the street due to the on-coming ego-vehicle. In \cref{fig:q_examp} (bottom left) our Joint-$\beta$-cVAE model correctly predicts that the pedestrian crosses the street and the ego-vehicle yields to the pedestrian. Similarly, in \cref{fig:q_examp} (bottom right) our Joint-$\beta$-cVAE model correctly predicts that both pedestrians cross the street (while avoiding collisions) and the the ego-vehicle yields to the pedestrians.
These examples show that our Joint-$\beta$-cVAE model can successfully capture the effect of interactions on the multi-modal distribution of pedestrian trajectories in the latent space.

\begin{figure*}[t]
  \centering
  \begin{tabularx}{\textwidth}{@{}ccc@{}}
	\toprule
	\textbf{On-board Observation} & \textbf{$L_2$ Norm of Velocity} & \textbf{$L_2$ Norm of Acceleration} \\
	\midrule
	\includegraphics[width=0.3\linewidth,height= 2.7 cm]{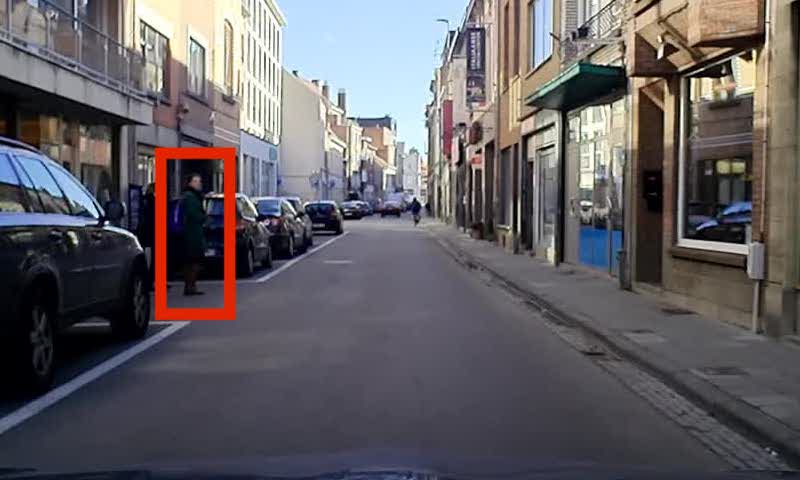}  &
	\includegraphics[width=0.32\linewidth,height= 2.7 cm]{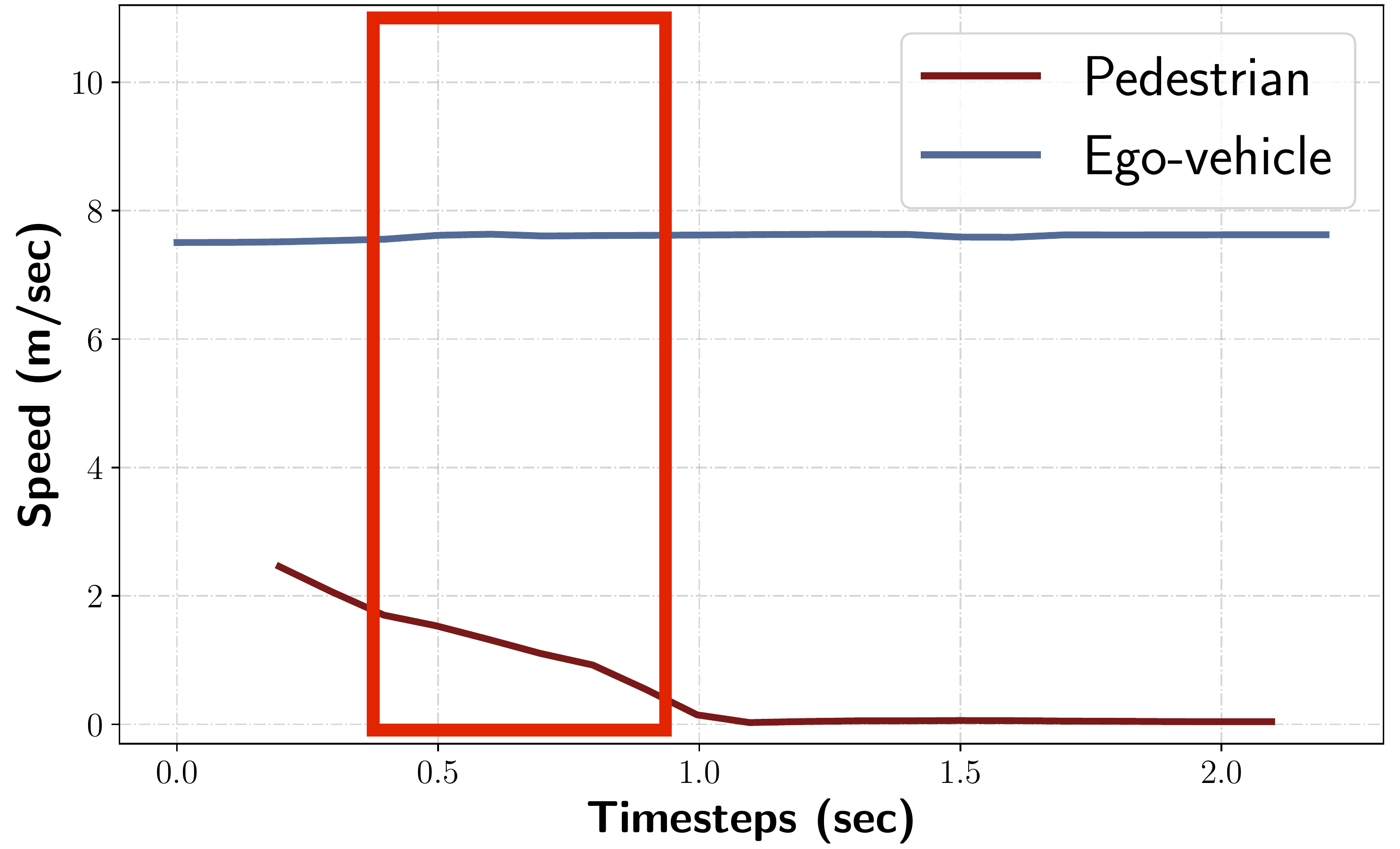} &
	\includegraphics[width=0.32\linewidth,height= 2.7 cm]{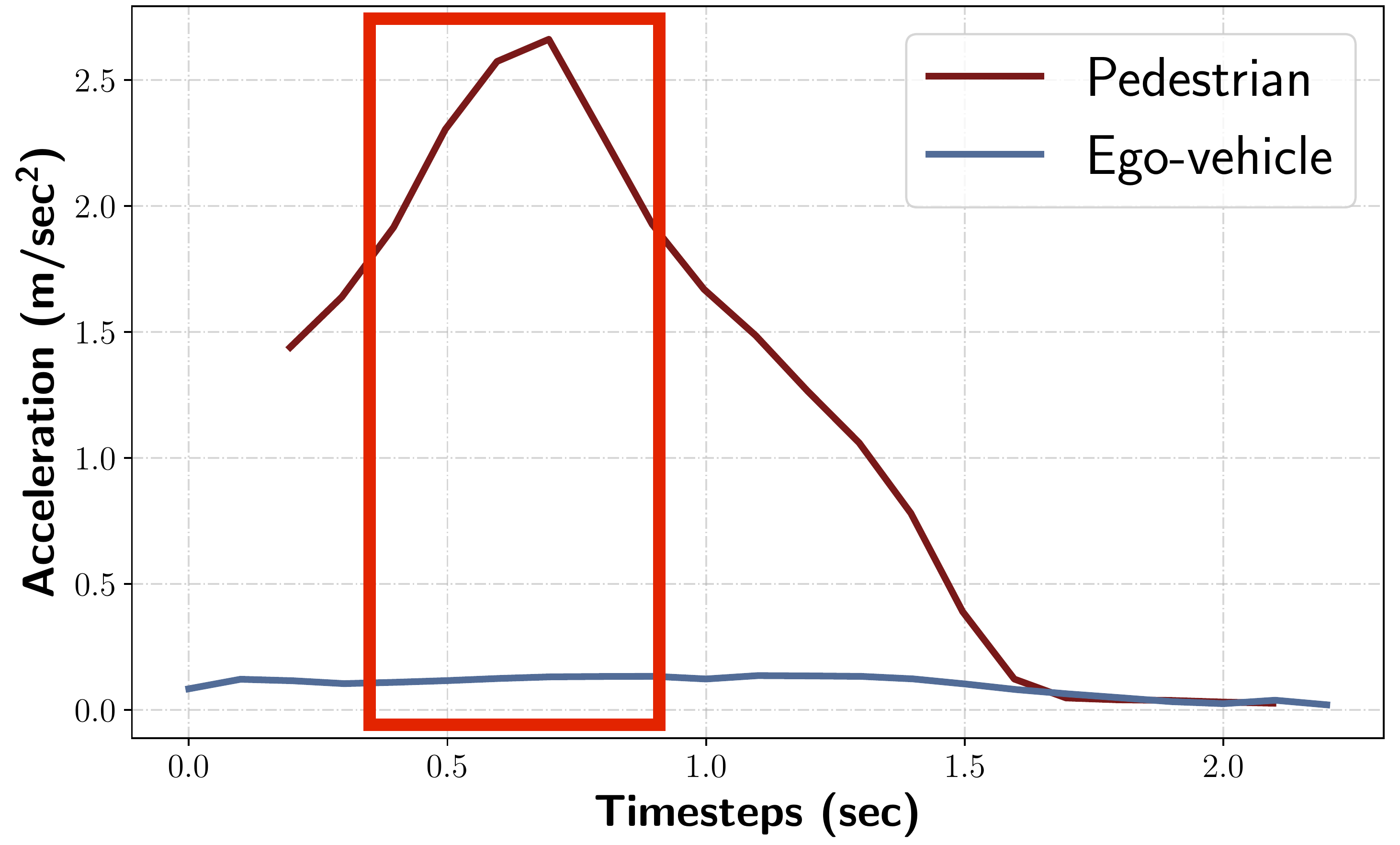}\\
	\cdashlinelr{1-3}
	\multicolumn{3}{c}{{\color[HTML]{791919} Pedestrian}: \textbf{Yields to vehicle as it has right of way.}  {\color[HTML]{526C97}Ego-vehicle}: \textbf{Continues at constant speed. }} \\
	\midrule
	\includegraphics[width=0.3\linewidth,height= 2.7 cm]{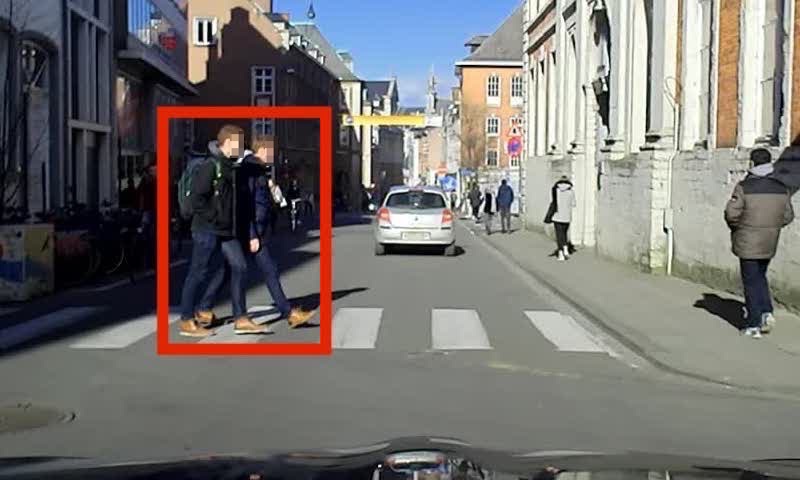}  &
	\includegraphics[width=0.32\linewidth,height= 2.7 cm]{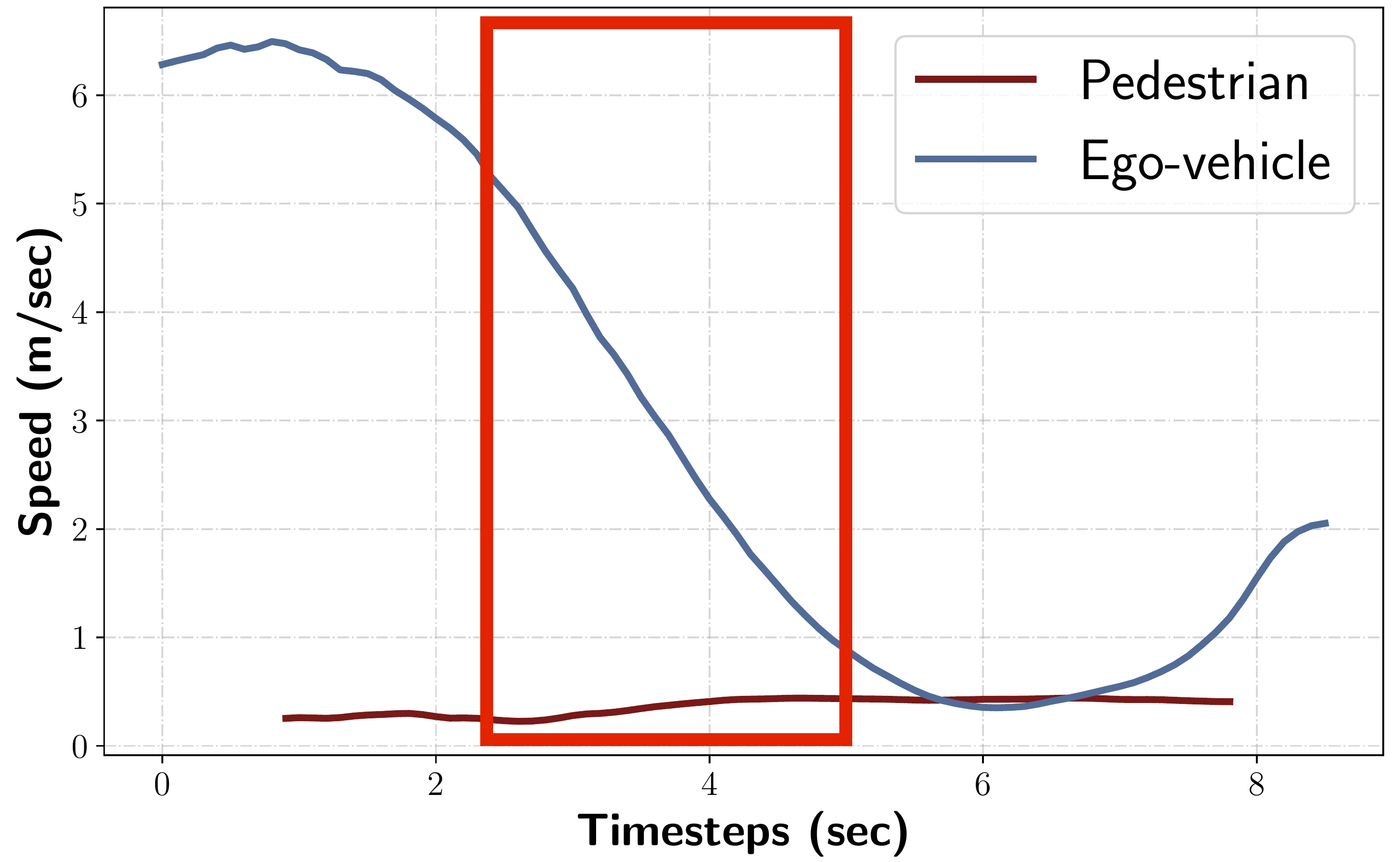} &
	\includegraphics[width=0.32\linewidth,height= 2.7 cm]{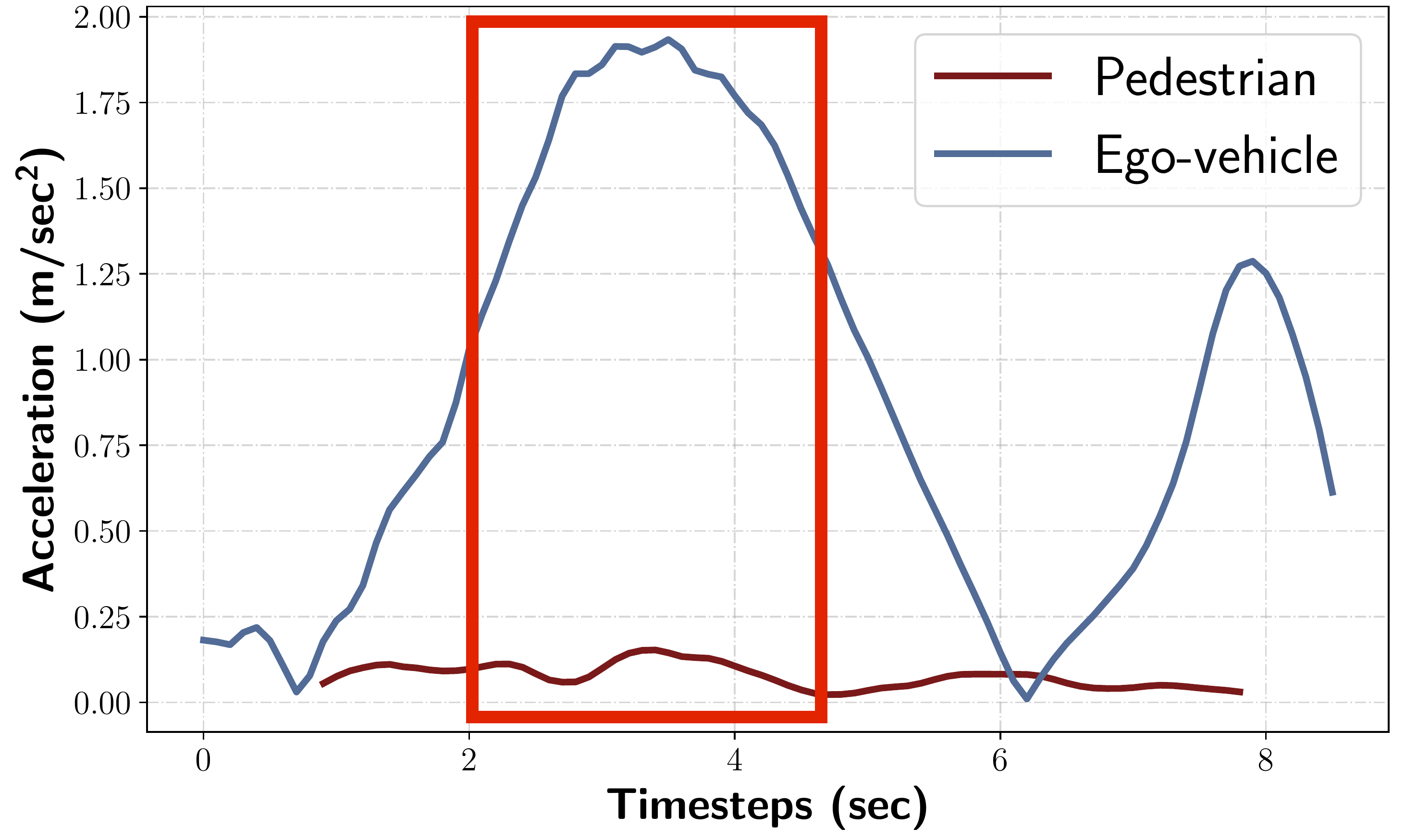}\\
	\cdashlinelr{1-3}
	\multicolumn{3}{c}{{\color[HTML]{791919} Pedestrian}: \textbf{Crosses at crosswalk with right of way..}  {\color[HTML]{526C97}Ego-vehicle}: \textbf{Yields to pedestrian. }} \\
	\midrule
	\includegraphics[width=0.3\linewidth,height= 2.7 cm]{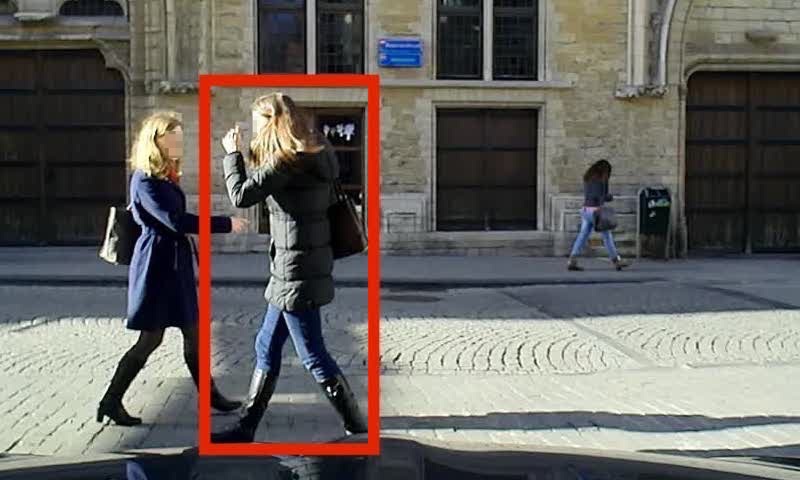}  &
	\includegraphics[width=0.32\linewidth,height= 2.7 cm]{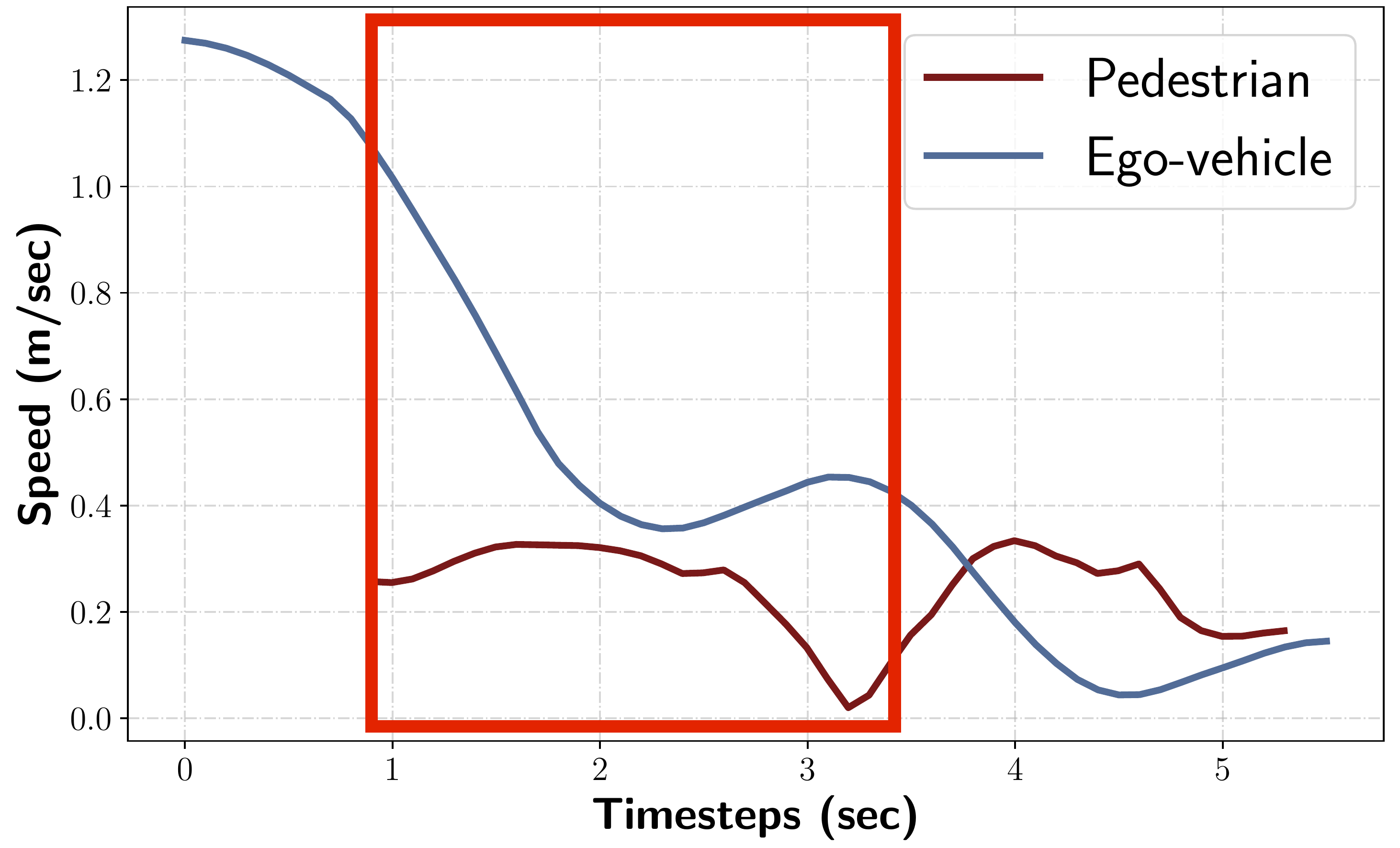} &
	\includegraphics[width=0.32\linewidth,height= 2.7 cm]{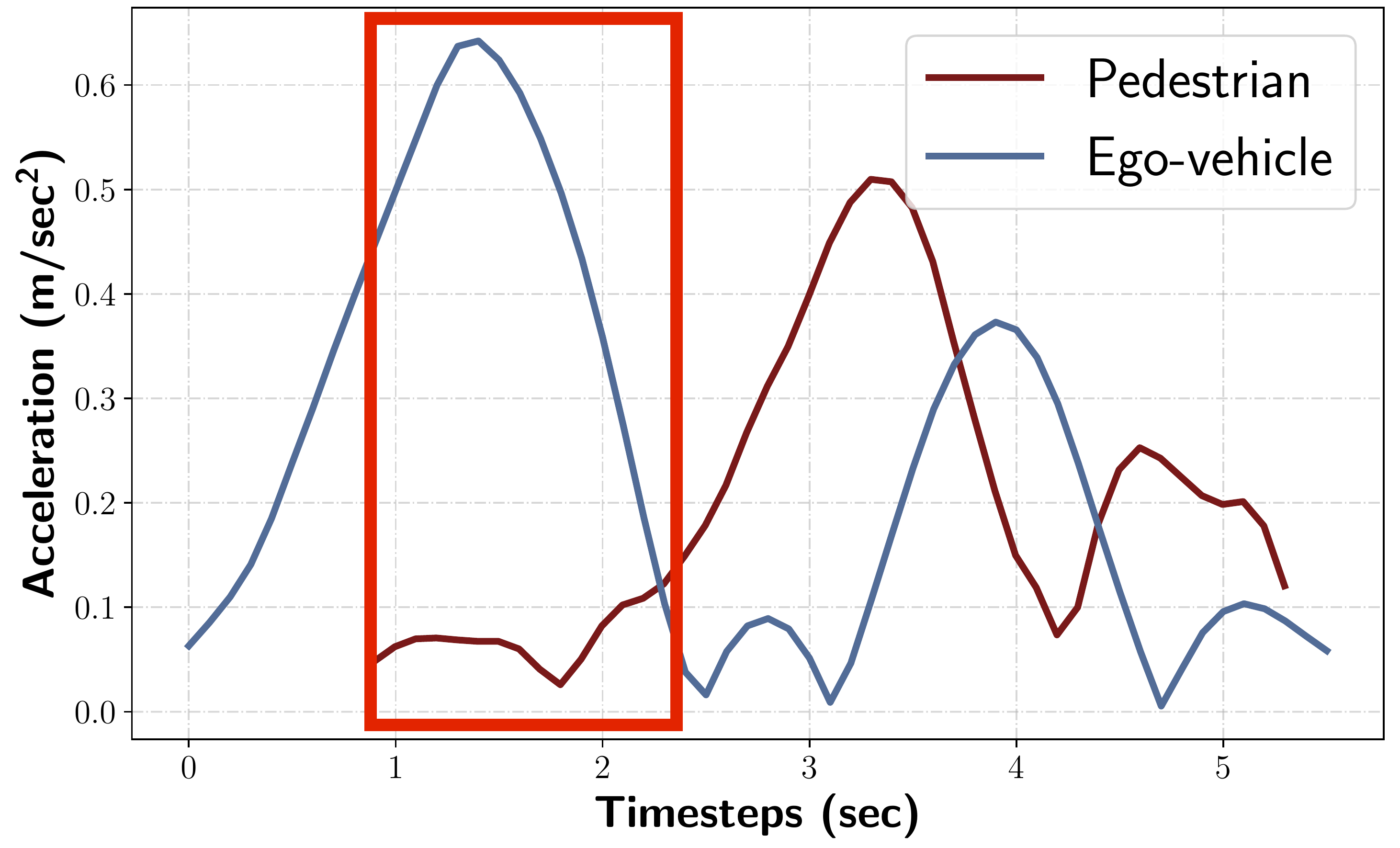}\\
	\cdashlinelr{1-3}
	\multicolumn{3}{c}{{\color[HTML]{791919} Pedestrian}: \textbf{Slows down after meeting acquaintance.}  {\color[HTML]{526C97}Ego-vehicle}: \textbf{Yields to pedestrian. }} \\
	\midrule
	\includegraphics[width=0.3\linewidth,height= 2.7 cm]{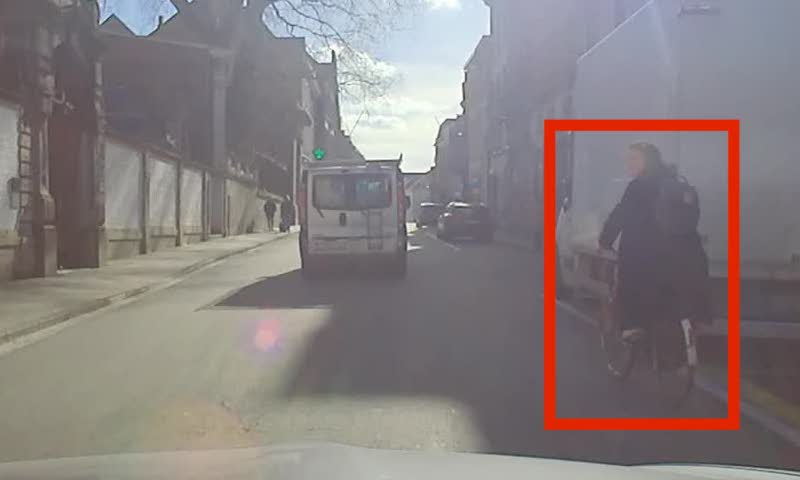}  &
	\includegraphics[width=0.32\linewidth,height= 2.7 cm]{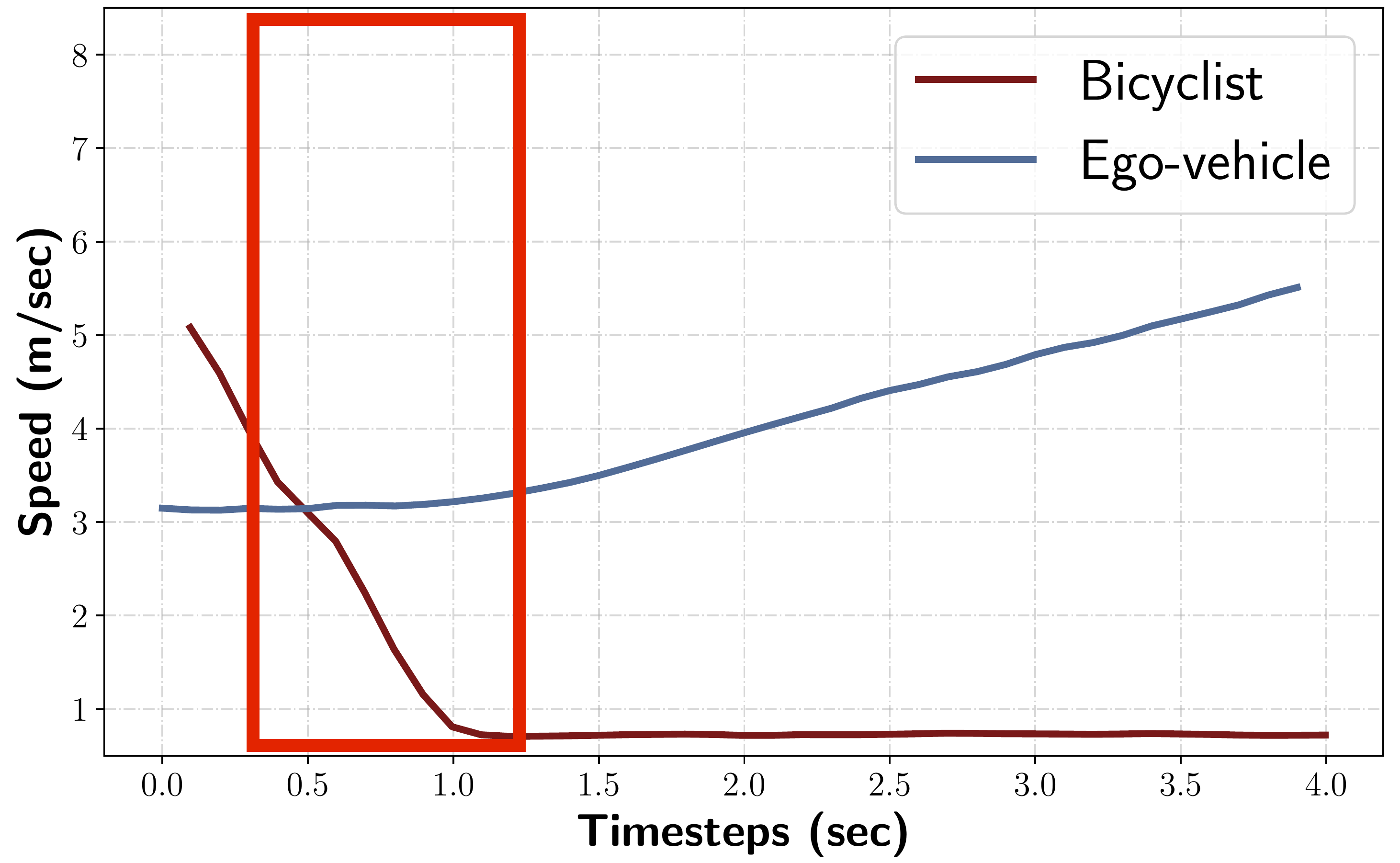} &
	\includegraphics[width=0.32\linewidth,height= 2.7 cm]{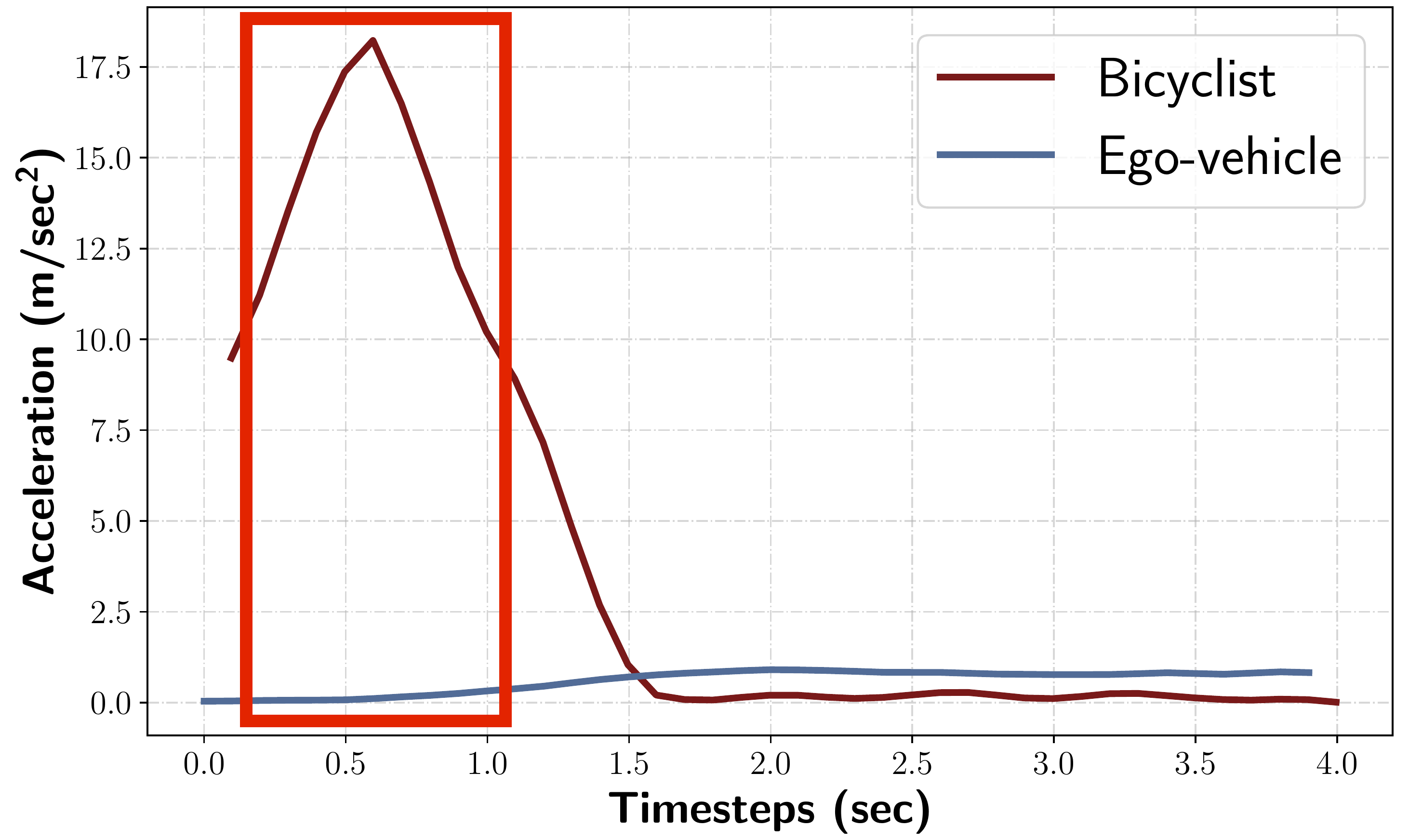}\\
	\cdashlinelr{1-3}
	\multicolumn{3}{c}{{\color[HTML]{791919} Bicyclist}: \textbf{Slows down due to approaching ego-vehicle in a narrow street.}  {\color[HTML]{526C97}Ego-vehicle}: \textbf{Continues at constant speed. }} \\
	\midrule
	\includegraphics[width=0.3\linewidth,height= 2.7 cm]{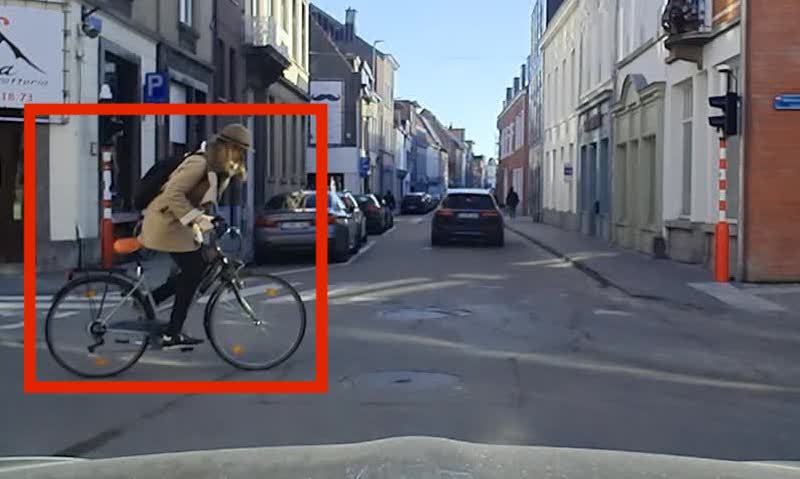}  &
	\includegraphics[width=0.32\linewidth,height= 2.7 cm]{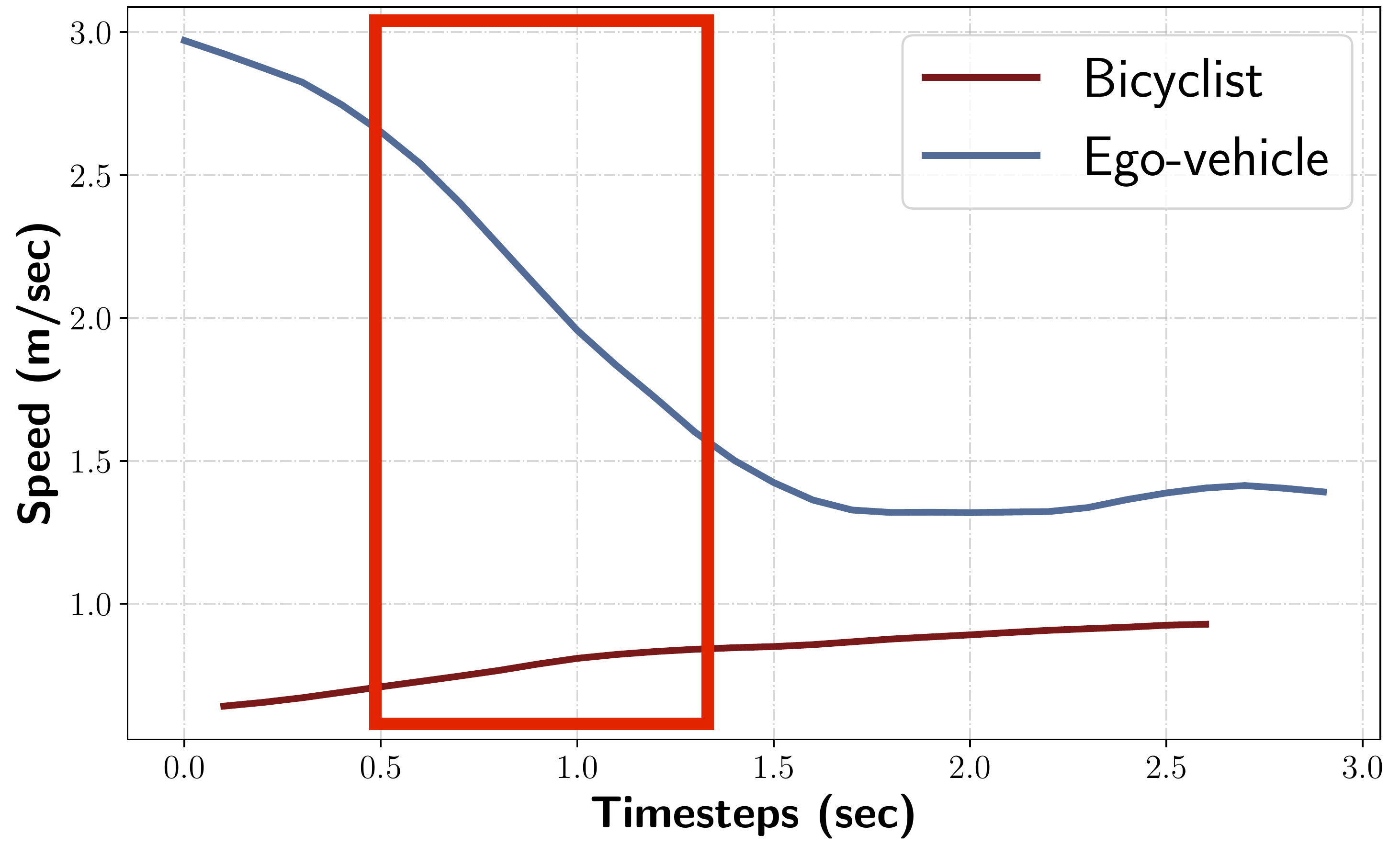} &
	\includegraphics[width=0.32\linewidth,height= 2.7 cm]{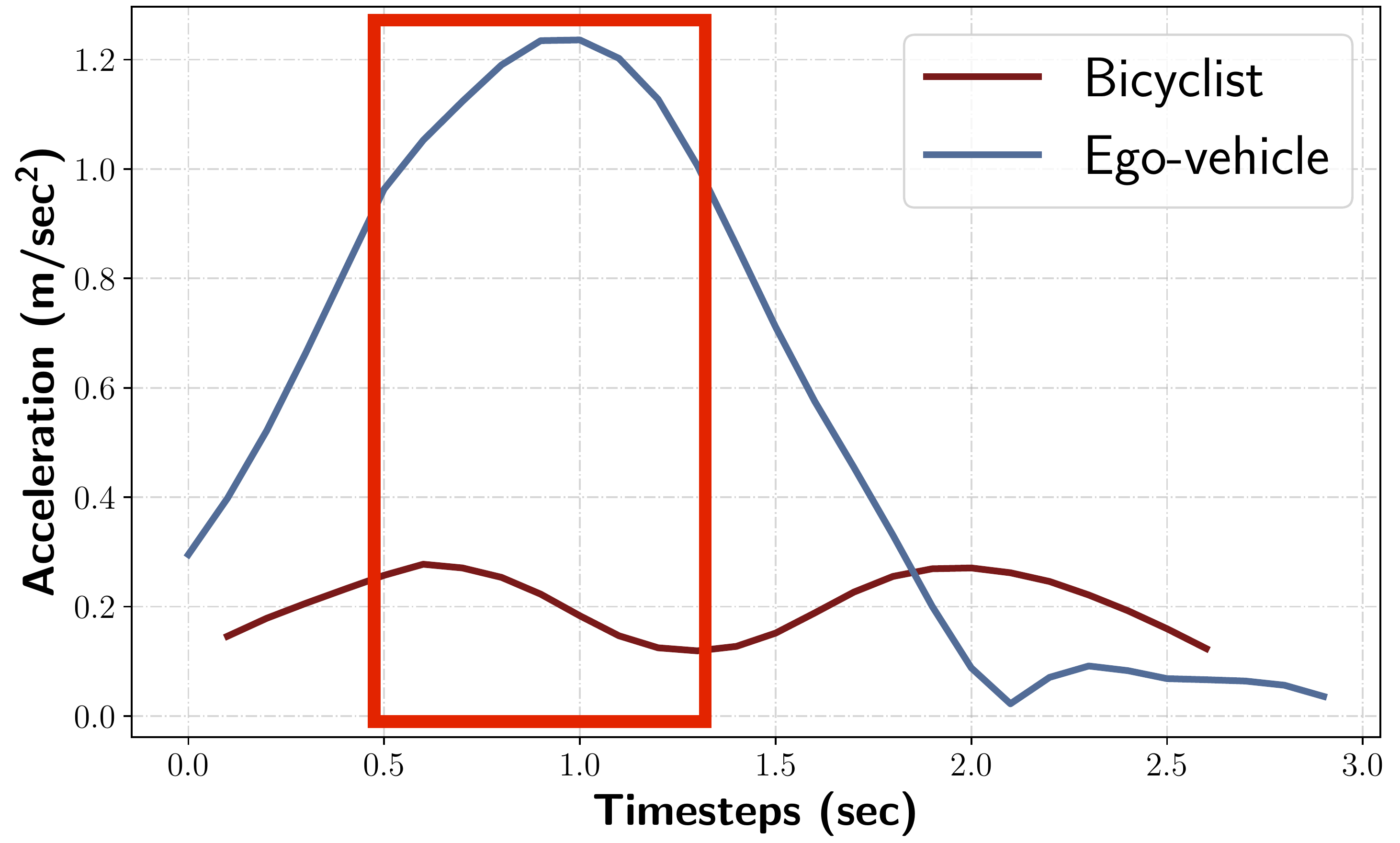}\\
	\cdashlinelr{1-3}
	\multicolumn{3}{c}{{\color[HTML]{791919} Bicyclist}: \textbf{Crosses in front of the ego-vehicle.}  {\color[HTML]{526C97}Ego-vehicle}: \textbf{Yields to bicyclist. }} \\
	\bottomrule
  \end{tabularx}
  \caption{Examples of interactions in the proposed Euro-PVI dataset. Spikes in the magnitude ($L_2$ norm) of acceleration resulting from interactions are marked. Top row: the pedestrian attempting to cross the road yields to the on-coming ego-vehicle as it has right of way. Second row: the ego-vehicle yields to the pedestrians as they are at a crosswalk and thus have the right of way. Third row: the pedestrian crossing the street in front of the ego-vehicle slows down after meeting an acquaintance and the ego-vehicle yields to the pedestrians and waits for them to cross the street. Fourth row: the bicyclist slows down to let the ego-vehicle pass due to the narrow street. Fifth row: the bicyclist crosses in front of the ego-vehicle and the ego-vehicle yields to the bicyclist. }
  \label{fig:example_interactions}
\end{figure*}

\begin{figure*}[t]
\centering
\begin{subfigure}[t]{.47\textwidth}
  \centering
  \frame{\includegraphics[height=4.75cm]{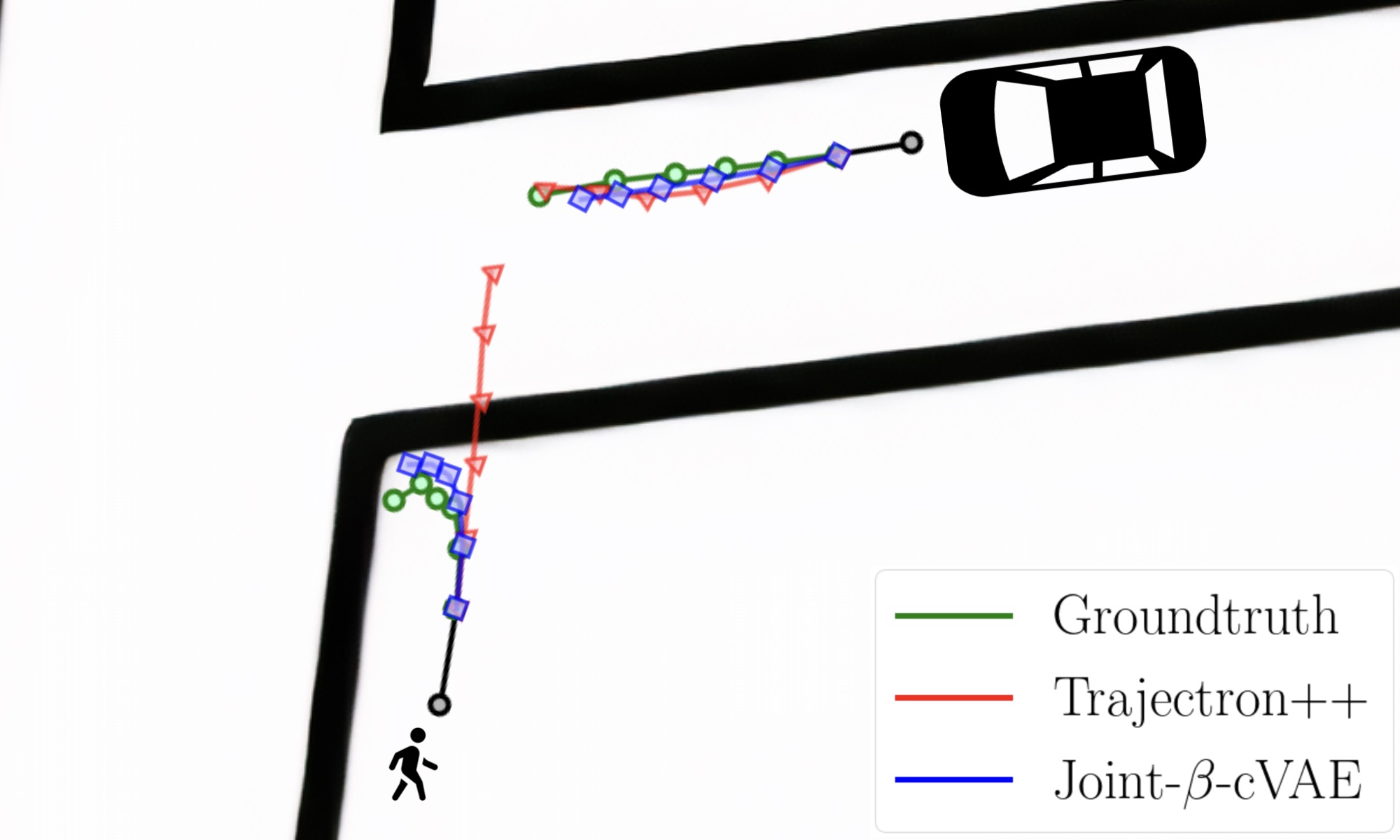}}
\end{subfigure}%
\hfill
\begin{subfigure}[t]{.47\textwidth}
  \centering
  \frame{\includegraphics[height=4.75cm]{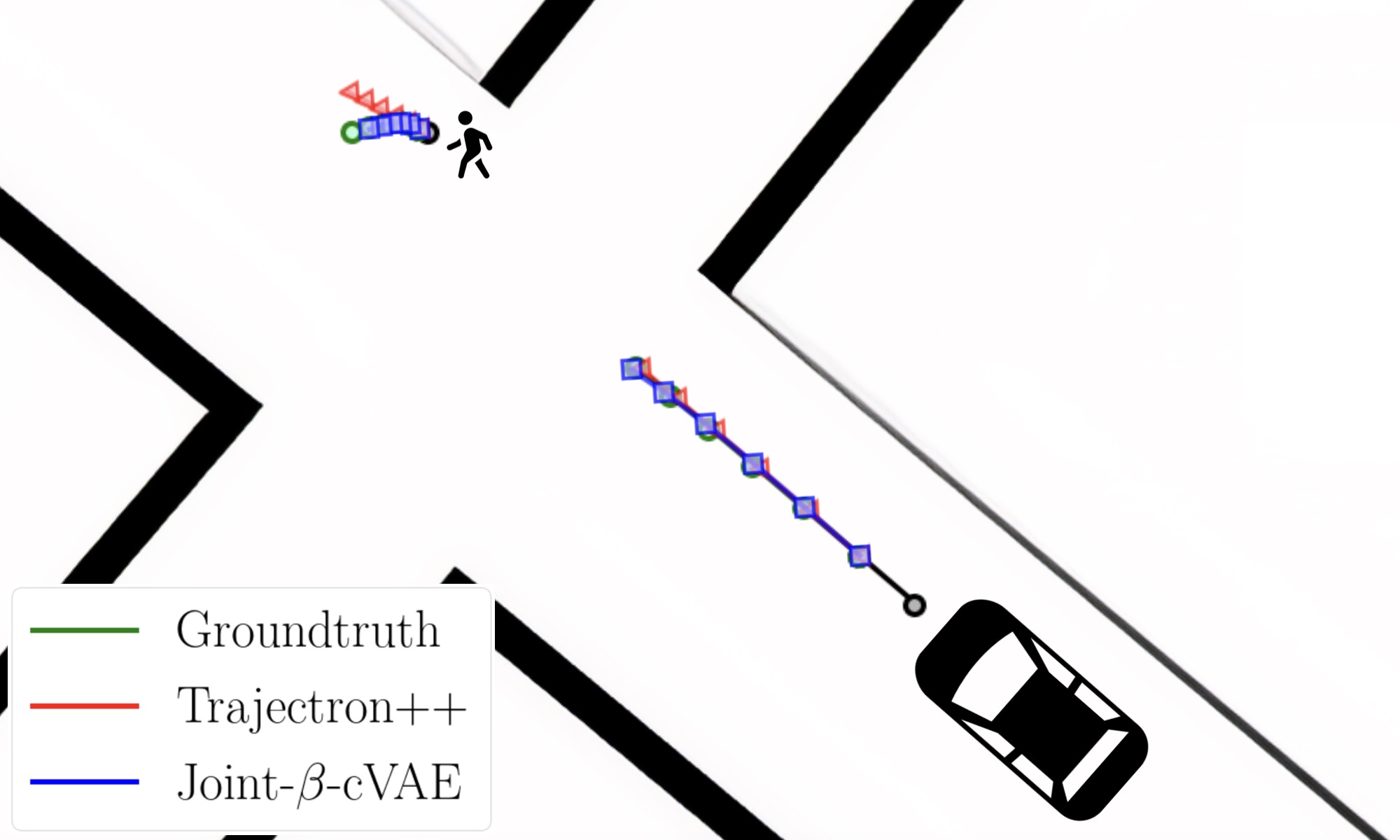}}
\end{subfigure}
\par\bigskip
\begin{subfigure}[t]{.47\textwidth}
  \centering
  \frame{\includegraphics[height=4.75cm]{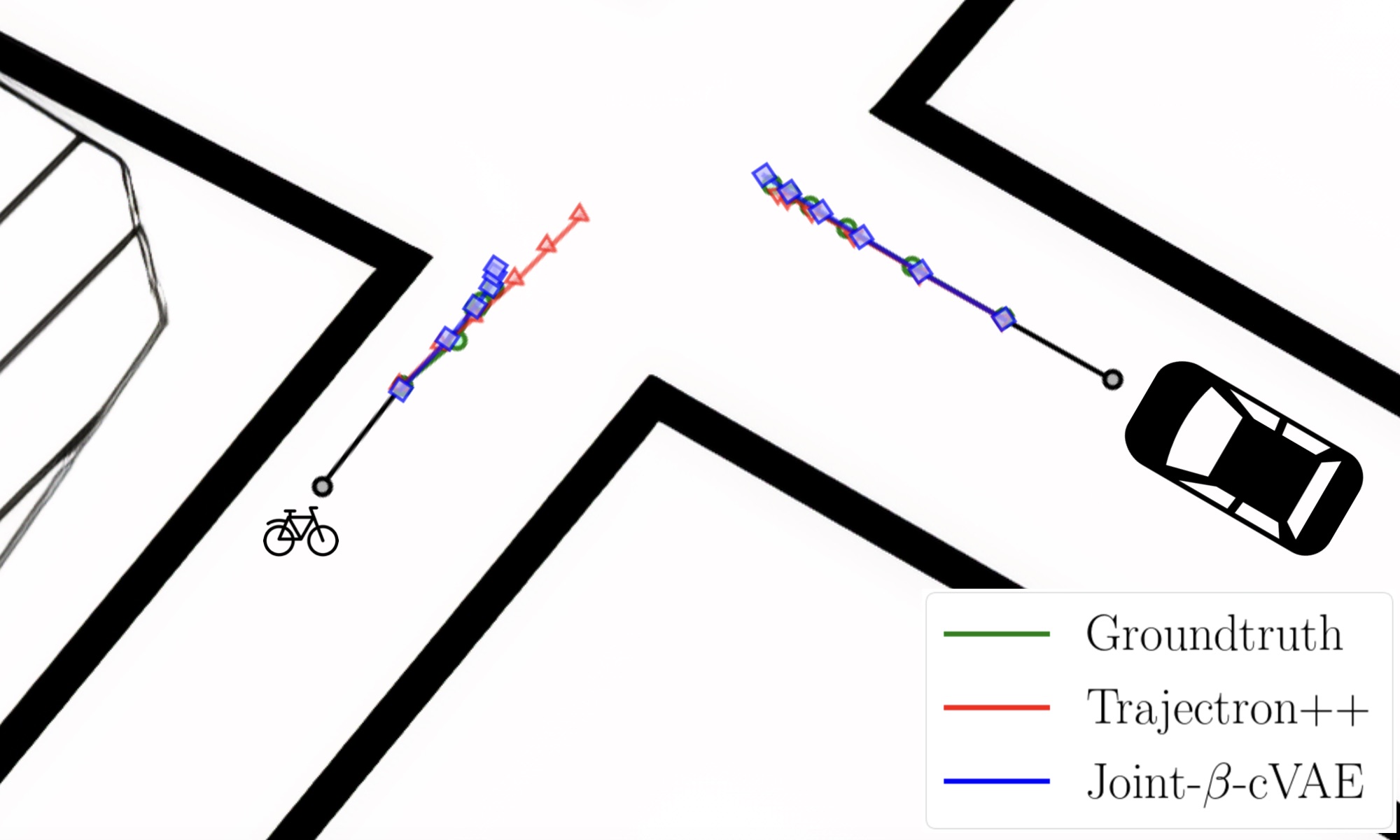}}
\end{subfigure}%
\hfill
\begin{subfigure}[t]{.47\textwidth}
  \centering
  \frame{\includegraphics[height=4.75cm]{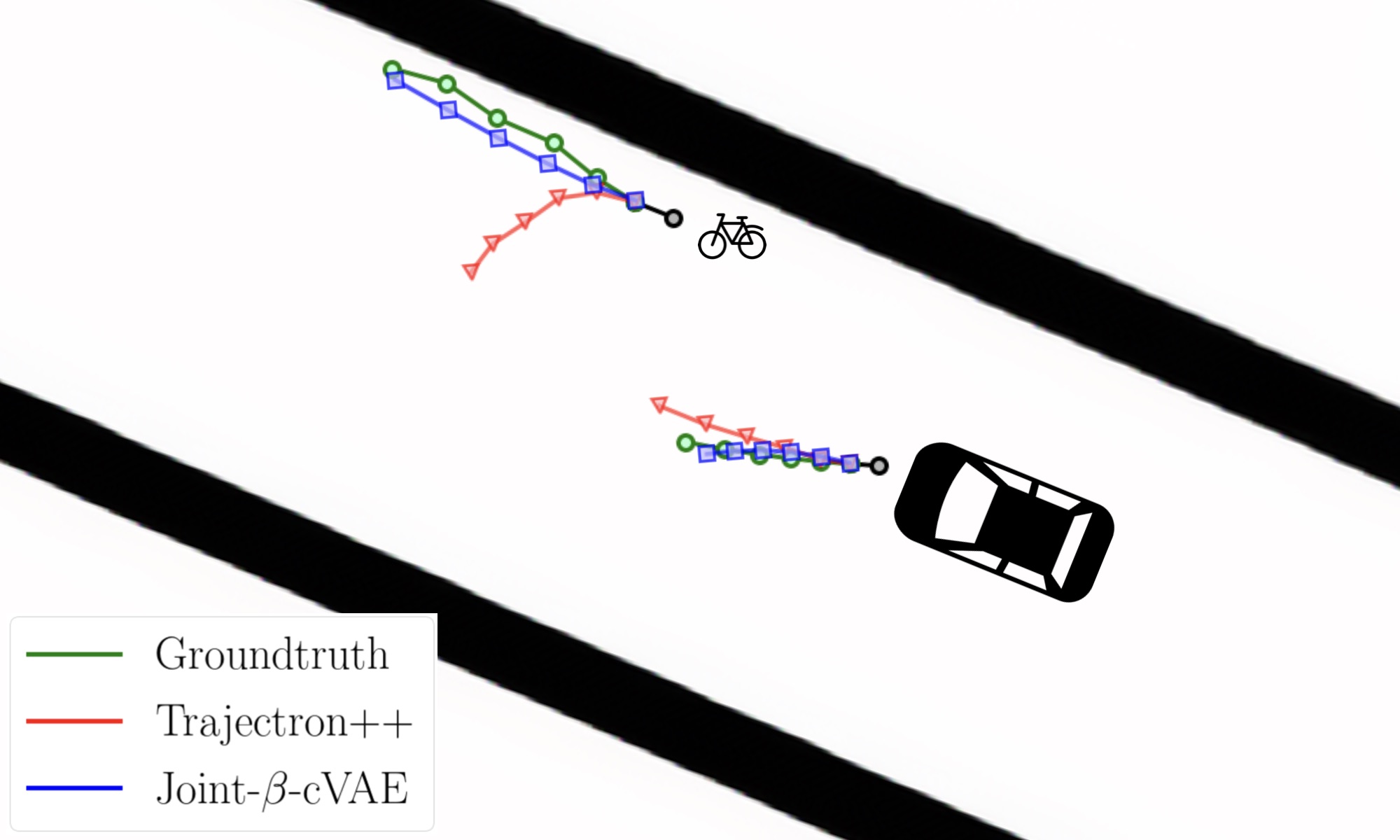}}
\end{subfigure}%
\par\bigskip
\begin{subfigure}[t]{.47\textwidth}
  \centering
  \frame{\includegraphics[height=4.75cm]{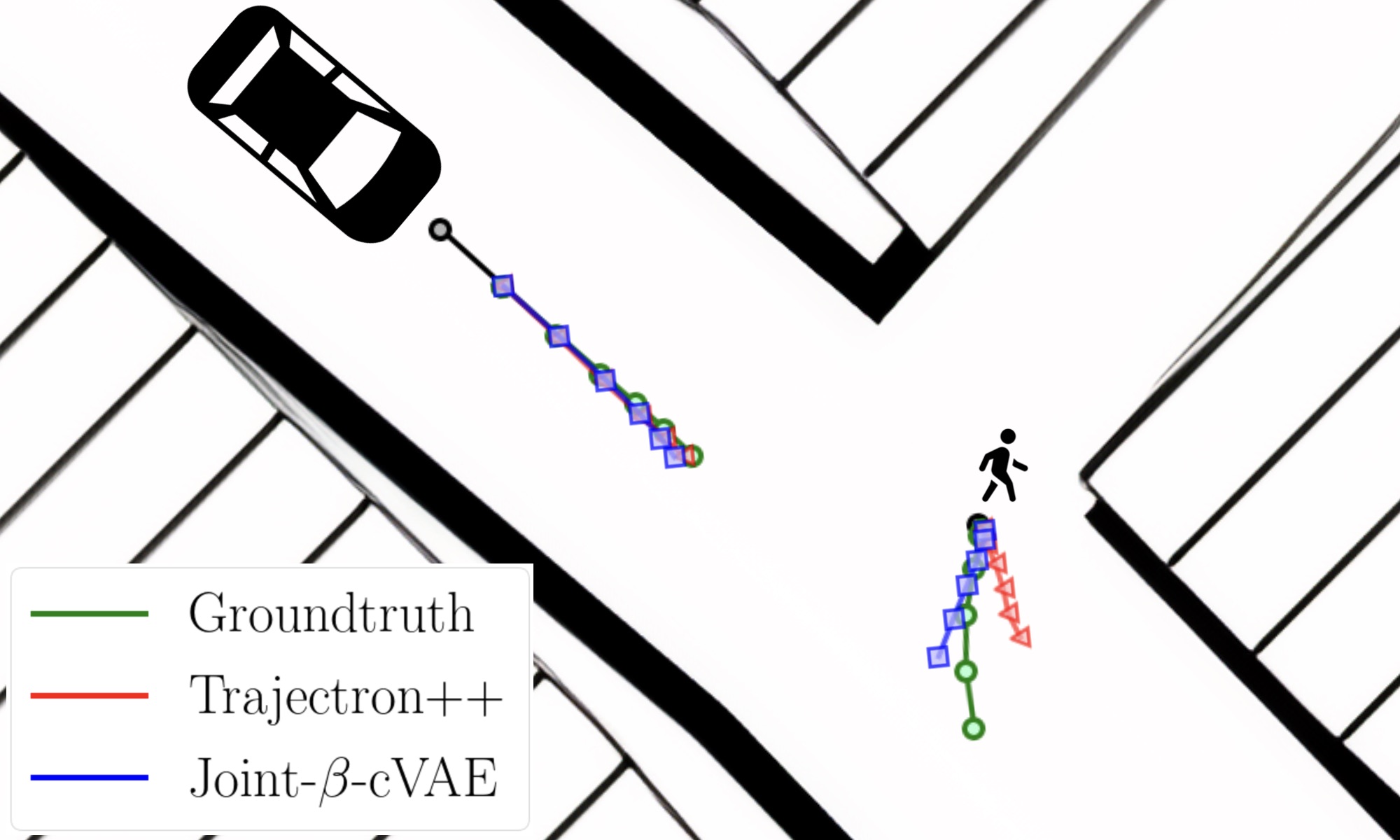}}
\end{subfigure}%
\hfill
\begin{subfigure}[t]{.47\textwidth}
  \centering
  \frame{\includegraphics[height=4.75cm]{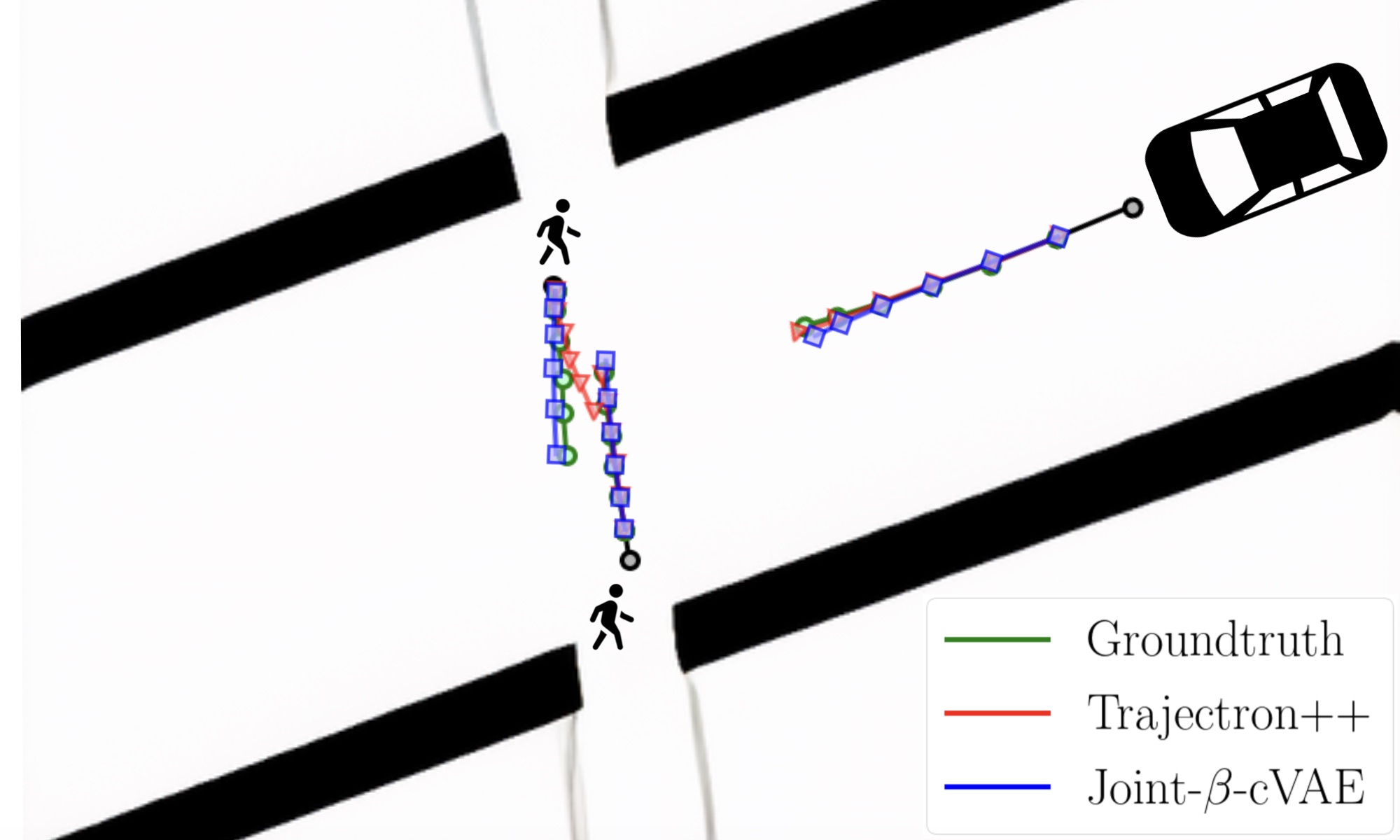}}
\end{subfigure}%

\caption{Qualitative examples on the Euro-PVI dataset. We compare the Best of $N\!=\!20$ samples for Trajectron++ (red) and our Joint-$\beta$-cVAE (blue).}
\label{fig:q_examp}
\end{figure*}

\end{document}